%% file: grendelxs-main.tex
\documentclass[sigplan,nonacm]{acmart}

\usepackage{hyperref}
\usepackage{url}

\usepackage{graphicx}
\usepackage{algorithm}
\usepackage{algpseudocode}
\usepackage{tabularx}
\usepackage[labelfont={small,it,bf},textfont={small,it}]{caption}
\usepackage{subcaption}
\usepackage{svg}
\usepackage{wrapfig}
\usepackage{multirow}
\usepackage[normalem]{ulem}
\usepackage[utf8]{inputenc} %
\usepackage[T1]{fontenc}    %
\usepackage{hyperref}       %
\usepackage{url}            %
\usepackage{booktabs}       %
\usepackage{amsfonts}       %
\usepackage{nicefrac}       %
\usepackage{microtype}      %
\usepackage{xcolor}         %
\usepackage{amsmath}
\usepackage{xspace}
\usepackage{paralist}
\usepackage{tikz}
\usepackage[compact]{titlesec}
\usepackage{tabularx}
\usepackage{soul} 

\newcommand{\name}[0]{CLM\xspace}

\newcommand{\comments}[1]{}
\renewcommand{\comments}[1]{#1}

\definecolor{darkgreen}{rgb}{0.0,0.5,0.0}
\newcommand{\revise}[1]{\textcolor{darkgreen}{#1}}
\renewcommand{\revise}[1]{#1}

\newcommand*\circled[1]{\tikz[baseline=(char.base)]{
            \node[shape=circle,fill,inner sep=0.8pt] (char) {\textcolor{white}{#1}};}}

\begin{document}
\pagestyle{plain}

\title{\name: Removing the GPU Memory Barrier for 3D Gaussian Splatting}
\thanks{This paper has been accepted to appear in the 2026 ACM International Conference on Architectural Support for Programming Languages and Operating Systems (ASPLOS~'26).}

\author{Hexu Zhao}
\authornote{Hexu and Xiwen contributed equally. }
\affiliation{
  \institution{New York University}
  \city{New York}
  \state{NY}
  \country{USA}
}
\email{hz3496@nyu.edu}

\author{Xiwen Min}
\authornotemark[1] %
\affiliation{
  \institution{New York University}
  \city{New York}
  \state{NY}
  \country{USA}
}
\email{xm2336@nyu.edu}

\author{Xiaoteng Liu}
\affiliation{
  \institution{New York University}
  \city{New York}
  \state{NY}
  \country{USA}
}
\email{xiaoteng.liu@nyu.edu}

\author{Moonjun Gong}
\affiliation{
  \institution{New York University}
  \city{New York}
  \state{NY}
  \country{USA}
}
\email{gw2396@nyu.edu}

\author{Yiming Li}
\affiliation{
  \institution{New York University}
  \city{New York}
  \state{NY}
  \country{USA}
}
\email{yl7516@nyu.edu}

\author{Ang Li}
\affiliation{
  \institution{Pacific Northwest National Laboratory \& University of Washington}
  \city{Seattle}
  \state{WA}
  \country{USA}
}
\email{ang.li@pnnl.gov}

\author{Saining Xie}
\affiliation{
  \institution{New York University}
  \city{New York}
  \state{NY}
  \country{USA}
}
\email{sx352@nyu.edu}

\author{Jinyang Li}
\affiliation{
  \institution{New York University}
  \city{New York}
  \state{NY}
  \country{USA}
}
\email{jinyang@cs.nyu.edu}

\author{Aurojit Panda}
\affiliation{
  \institution{New York University}
  \city{New York}
  \state{NY}
  \country{USA}
}
\email{apanda@cs.nyu.edu}

\begin{abstract}
3D Gaussian Splatting (3DGS) is an increasingly popular novel view synthesis approach due to its fast rendering time, and high-quality output. However, scaling 3DGS to large (or intricate) scenes is challenging due to its substantial memory requirement, which exceeds the memory capacity of most GPUs. In this paper, we describe \name, a system that allows 3DGS to render large scenes using a single consumer-grade GPU, e.g., RTX4090. It does so by offloading Gaussians to CPU memory, and loading them into GPU memory only when necessary. To improve performance and reduce communication overheads, \name uses a novel offloading strategy based on insights into 3DGS's memory access patterns. This strategy enables efficient pipelining, which overlaps GPU-to-CPU communication, GPU computation and CPU computation.  Furthermore, \name exploits these access patterns to reduce communication volume. Our evaluation shows that the resulting implementation can render a large scene that requires 102 million Gaussians on a single RTX4090 and achieve state-of-the-art reconstruction quality.

\end{abstract}

\maketitle %

\input{01_intro}

\input{02_background_revision}

\input{03_design_revision}

\input{04_eval_revision}

\input{05_related_revision}

\input{06_conclusion_revision}

\bibliographystyle{ACM-Reference-Format}
\bibliography{grendelxs-bib}
\appendix
\input{07_appendix_revision}

\end{document}

%% file: 01_intro.tex
\section{Introduction}\label{sec:introduction}
Recently, there has been significant interest~\cite{3dgs, grendel, reduce3dgs,survey,ali2025compression3dgssurvey,bao2025survey3dgs} in using 3D Gaussian Splatting (3DGS) for novel view synthesis (Figure~\ref{fig:nvs}). Given a set of \emph{posed images} (i.e., images with position and orientation) for a 3D scene, 3DGS iteratively trains a \emph{scene representation} that consists of a large number of anisotropic 3D Gaussians that capture the scene's appearance and geometry. Users can then use the trained scene representation to render images from a previously unseen view. Compared to other novel view synthesis approaches, 3DGS has faster rendering time while achieving comparable image quality, thus leading to its surging popularity.

The quality of images rendered using 3DGS depends on the fidelity of the trained scene representation. Scenes that capture a large area or contain intricate details require a larger number of Gaussians. As a result, 3DGS's memory footprint grows as scene size, scene intricacy, or output image resolution increases. State-of-the-art 3DGS implementations run on GPUs, where memory is not plentiful. Therefore, memory capacity has been a barrier when scaling 3DGS and applying it to large intricate scenes with high image resolution. As we explain in \S\ref{sec:related}, prior work on scaling 3DGS either adds significant cost because they use multiple GPUs~\cite{grendel,dogs, retinags}, or compromises rendering quality because they reduce the scene representation's fidelity~\cite{lp3dgs,reduce3dgs,lightgaussian, minisplatting,citygaussian,vastgaussian,hierarchicalgaussians}.

In this paper, we describe \name, a system that scales 3DGS without requiring multiple GPUs or hurting rendered image quality. \name's design is based on the insight that the computation of 3DGS is inherently sparse; i.e. each training iteration only accesses a small subset of the scene's Gaussians.  Thus, it is sufficient to load only this subset into GPU memory, while leaving the remaining Gaussians offloaded to the more plentiful CPU memory. Despite this straightforward insight, as the GPU-CPU communication incurs significant overhead, it is challenging to realize the idea of memory offloading with good performance. %

We develop a novel 3DGS-specific offloading strategy for \name. Our offloading strategy minimizes performance overheads and scales to large scenes by leveraging four observations (\S\ref{sec:opportunity:frustum-culling}) about the 3DGS training pipeline: 
\begin{compactenum}[(i)]
    \item The set of Gaussians accessed by each view (aka a training image) can be computed ahead-of-time, thereby allowing the loading of Gaussians for one iteration to be overlapped with the computation for the previous iteration to reduce communication overhead (\S\ref{sec:design:attributewise-offload}, \S\ref{sec:design:microbatches-pipeline}).
    \item There is substantial overlap between the Gaussians accessed by different views, which allows us to cache the overlapping Gaussians to reduce the communication volume during each training iteration (\S\ref{sec:design:gaussian-caching}).
    \item The training process exhibits spatial locality, i.e., views in the same region tend to access the same Gaussians, allowing us to schedule training iterations carefully to maximize overlapped accesses across successive iterations in order to minimize overall communication volume (\S\ref{sec:design:order-optimization}).
    \item We can further use spatial locality to overlap gradient computation and a substantial portion of the Gaussian parameter update (\S\ref{sec:design:overlap-adam}).
\end{compactenum}

By exploiting the inherent sparsity in scenes and the four observations above, \name can scale to very large scenes: our evaluation (\S\ref{sec:eval}) shows that we can train a large scene with 100 million Gaussians on an a consumer-grade GPU (RTX4090) while achieving output quality on par with (or better than) the state-of-the-art systems. %
Furthermore, we show that \name's 3DGS-specific offloading solution incurs modest performance overheads compared to a baseline without offloading when rendering small scenes that can fit in the baseline system's GPU memory.

The rest of this paper is organized as follows: in \S\ref{sec:background} we provide background on novel-view synthesis and 3DGS; in \S\ref{sec:opportunity:frustum-culling} we detail observations about 3DGS's memory access patterns; in \S\ref{sec:design} we describe \name's design and in \S\ref{sec:impl} provide details about our implementation; we evaluate \name in  \S\ref{sec:eval}; and present related work in \S\ref{sec:related}.

%% file: 02_background_revision.tex
\section{Background and Motivation}\label{sec:background}

This section gives an overview of the 3DGS algorithm and its application, discusses its memory bottleneck, and explains the challenges associated with naive offloading strategies.

\subsection{Novel View Synthesis and 3D Gaussian Splatting}
\label{bg:3dgs} 

\begin{figure}[t]
    \centering
    \includegraphics[width=1\linewidth]{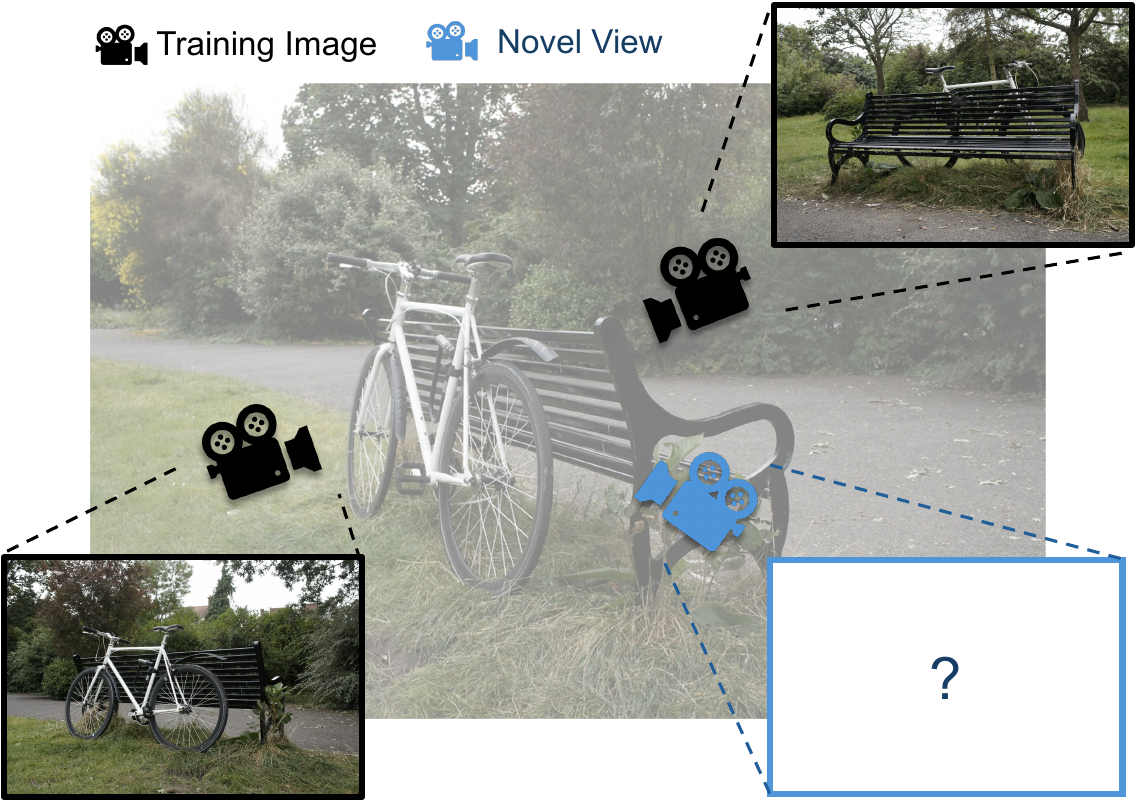}
    \caption{The novel view synthesis problem: given a set of training images (with known pose) from a scene,  render the image from a novel view with an unrecorded camera position and orientation. 
    }
    \label{fig:nvs}
\end{figure}

Novel View Synthesis (NVS) is the task of rendering an image of a 3D scene from a previously unseen viewpoint. To do so, NVS algorithms take as input a set of posed images (Figure~\ref{fig:nvs}), and use this input to construct a \emph{scene representation}, i.e., a 3D model that captures the scene's appearance and geometry. The scene representation is then used to render the desired image. Modern ML-based NVS approaches all aim to learn how to reconstruct the scene from input data but differ in how the scene is represented: 3DGS uses 3D Gaussians~\cite{3dgs} while others have used a mesh~\cite{softrasterizer} or a neural network~\cite{nerf}.

\begin{figure}[t]
    \centering
    \includegraphics[width=1\linewidth]{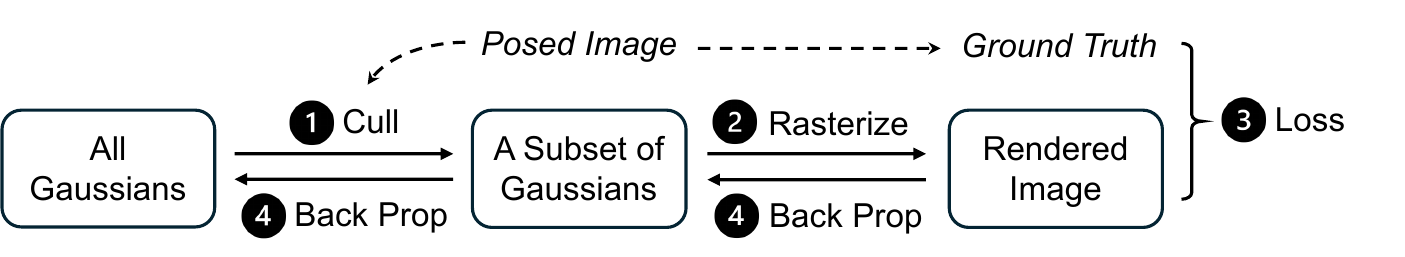}
    \caption{3D Gaussian Splatting Illustration. 
    }
    \label{fig:3dgs-algo}
\end{figure}

\begin{table}[h]
    \centering
    \resizebox{\columnwidth}{!}{
\begin{tabular}{lcccc|c}
\toprule
\multirow{2}{*}{} & \multirow{2}{*}{\textbf{3D Position}} & \multirow{2}{*}{\begin{tabular}[c]{@{}c@{}}\textbf{Covariance}\\ \textbf{(Scale+Rotation)}\end{tabular}} & \multirow{2}{*}{\begin{tabular}[c]{@{}c@{}}\textbf{Spherical Harmonics}\\ \textbf{(Color)}\end{tabular}} & \multirow{2}{*}{\textbf{Opacity}} & \multirow{2}{*}{\textbf{Total}} \\
                  &                              &                                                                                        &                                                                                        &                          &                        \\ \midrule
\#Param             & \multicolumn{1}{c}{3}        & \multicolumn{1}{c}{3+4}                                                                & \multicolumn{1}{c}{48}                                                                 & \multicolumn{1}{c}{1}    & \multicolumn{1}{|c}{59} \\ \bottomrule
\end{tabular}
    }
    \caption{A 3D Gaussian has 59 learnable parameters representing 4 types of attributes. }
    \label{tab:3dgs_attributes}
\end{table}

3DGS represents the scene as a (potentially very large) collection of anisotropic 3D Gaussians, each of which consists of several dozen parameters representing four types of attributes including position, anisotropic covariance, spherical harmonic coefficients and opacity, as seen in Table~\ref{tab:3dgs_attributes}. These learnable parameters of a 3D Gaussian dictate its effects on a rendered scene.

3DGS' differentiable rendering allows it to use gradient-based optimization based on minibatch SGD. 
In particular, Gaussians are initialized either randomly or using a user-provided point cloud generated by COLMAP~\cite{colmap}. Then, training proceeds iteratively. At each training step, one (or a batch of) camera view is selected from among the views represented in the training data set. The selected view is rendered (Figure~\ref{fig:3dgs-algo}) by \circled{1} first selecting the set of Gaussians in the camera's frustum (referred to as frustum culling) and then \circled{2} rasterizing them. Afterwards, \circled{3} loss is computed by comparing the rendered image with the groundtruth training image, and \circled{4} backpropagation is run to update the Gaussian attributes. Periodically, adaptive densification~\cite{3dgs, hierarchicalgaussians, mcmc3dgs} is performed to increase the number of Gaussians in areas with high reconstruction errors and to prune unnecessary Gaussians. %

\subsection{Challenges of training 3DGS on a consumer GPU} 
\label{bg:challenges-of-training-3dgs-on-a-consumer-gpu}

\begin{table}[t]
    \centering
    \resizebox{\columnwidth}{!}{
        \begin{tabular}{lrrr}
            \toprule
            \textbf{Scene} & \textbf{Resolution} & \textbf{\# Gaussians} & \textbf{Memory Demand} \\
            \midrule
            Bicycle \cite{mipnerf360} & 4K & 9 M & 10 GB \\
            Rubble \cite{meganerf} & 4K & 40 M & 50 GB \\
            Alameda \cite{zipnerf} & 2K & 45 M & 60 GB \\
            Ithaca \cite{ithaca365} & 1K & 70 M & 80 GB \\
            Matrixcity BigCity \cite{matrixcity} & 1080P & 100 M & 110 GB \\
            \bottomrule
        \end{tabular}
    }
    \caption{This table details the necessary number of Gaussians and minimum memory requirements during 3DGS training for 3 scenes with varying resolution and Gaussians quantity. The Rubble, Alameda, Ithaca and BigCity datasets are much larger than the Bicycle dataset and demand more memory than a single 24GB RTX 4090 can supply. } 
    \label{tab:scene-aera-resolution-n3dgs-memory} 
\end{table}

\noindent\textbf{The memory barrier to scaling 3DGS.}
State-of-the-art 3DGS implementations~\cite{3dgs, gsplat,grendel} run on GPUs for performance. However, representing a complex scene using 3D Gaussians requires a significant amount of memory, more than what is available on most GPUs.
To estimate 3DGS' memory consumption, we observe that the model states (i.e., the set of Gaussians representing the scene) consume a majority of the memory used during rendering. As shown in Table~\ref{tab:3dgs_attributes}, each Gaussian has $59$ parameters, each of which results in four 4-byte floating point numbers stored during training: the parameter itself, its gradient, and two additional versions as the optimizer state~\cite{adam,3dgs}.
Thus, for a scene with $N$ Gaussians, the model state alone requires $N \times 59 \times 4 \times 4$ bytes, which can easily exceed the memory capacity of consumer-grade GPUs.  For example, RTX 4090 with 24GB can only store the memory state of up to 26 million Gaussians, even if we ignore the memory consumption of the activation state and various other temporary buffers.  Table~\ref{tab:scene-aera-resolution-n3dgs-memory} lists the number of Gaussians required to achieve good rendering quality for well-known NVS datasets. Except for the smallest scene (Bicycle), all other larger and more complex scenes such as Rubble, Alameda, Ithaca, and MatrixCity BigCity cannot be trained on a single consumer-grade GPU such as RTX 4090.

\begin{figure}[t]
    \centering
    \includegraphics[width=1\linewidth]{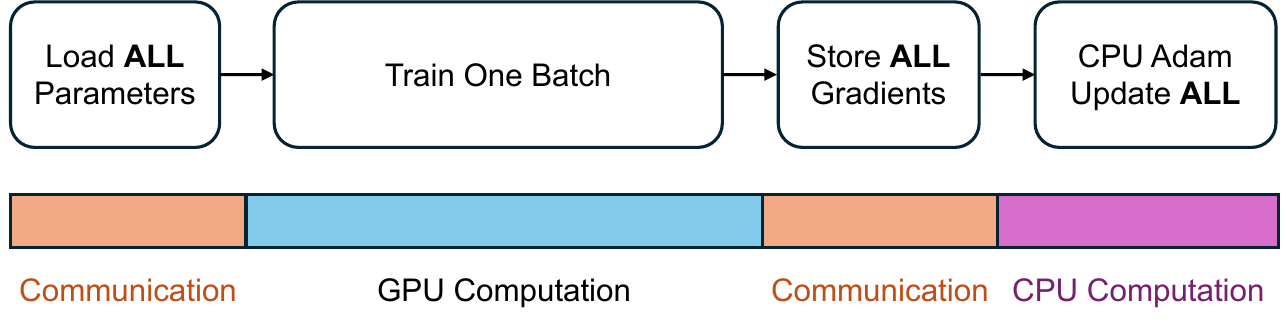}
    \caption{Runtime decomposition of one batch in naive offloading. It leads to overheads in communication and CPU Adam computation. }
    \label{fig:offload-simple}
\end{figure}

\noindent\textbf{Challenges of offloading. }
\label{sec:naive-offload}
Existing work addresses the memory barrier of 3DGS through multi-GPU training \revise{or Gaussian pruning}. \revise{Both approaches have shortcomings: using multiple GPU training significantly raises training cost, while Gaussian pruning not only degrades quality (see \S\ref{sec:related} for more details) but also fails to handle very large scenes, where even a pruned model may exceed GPU memory capacity.  In contrast, our work develops an orthogonal approach by offloading Gaussians to CPU memory. }

\revise{A simple approach to address this problem would be to use a technology such as Unified Virtual Memory that uses CPU memory to augment GPU memory, and swaps data in from main memory when required by the GPU. While this is indeed simple it has significant overheads~\cite{perfevalcudaunifiedmemory}, and would thus not suffice for our use case.}

\revise{
Work on deep-learning, e.g. Zero-Offload~\cite{zerooffload}, has shown that it is possible to train a large model without impacting quality by offloading the gradients, optimizer states and optimizer computation to CPU. Thus, one can ask  whether a similar offloading approach would work for 3DGS.}
Figure~\ref{fig:offload-simple} shows how such a Zero-offload inspired approach could work for 3DGS:
in each training step, first, all Gaussians are transferred from CPU memory to GPU memory; next, the forward and backward computation are carried out on the GPU; and finally, gradients are sent back to the CPU where the Adam optimizer~\cite{kingma2015adam} is run to update Gaussian parameters.  

\revise{However, naively applying Zero-offload leads to two problems: First, as Figure~\ref{fig:offload-simple} shows, naive offloading incurs significant performance overhead due to CPU-GPU communication and the additional time required to run Adam on the CPU. 
However, naive offloading lacks the means to effectively hide such overhead by overlapping GPU computation with communication and CPU computation. Second, as naive offloading loads all Gaussians to the GPU, its GPU memory requirement remains proportional to the number of Gaussian such that large scenes cannot fit on a single GPU.}

\section{Our approach: sparsity-guided offloading} %
\label{sec:opportunity:frustum-culling} 

\begin{figure}[t]
    \centering
    \includegraphics[width=0.6\linewidth]{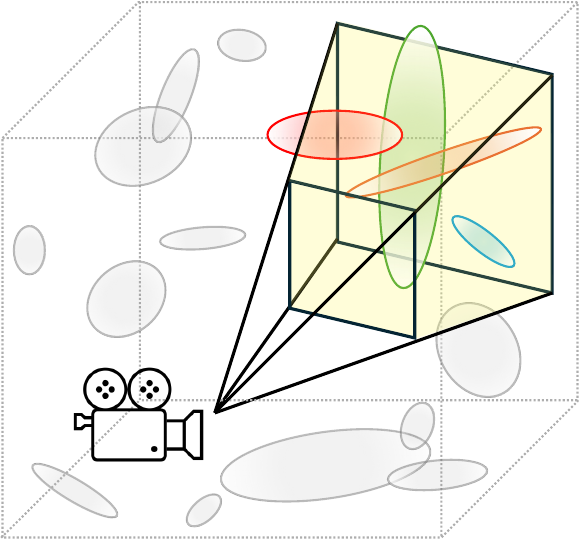}
    \caption{Frustum Culling: Gaussians outside of camera frustum will not be accessed when rendering the camera's view.
    Further, the Gaussians accessed when rendering a view are in the same region, i.e., the process exhibits spatial locality. Our approach uses these observations to improve performance and reduce GPU memory requirements.
     This results in a sparse memory access pattern to gaussians parameters. This also shows that 3DGS rendering has the property of spatial locality. 
     } 
    \label{fig:frustum-cull} 
\end{figure}

\revise{Our approach addresses the challenges we discussed above by storing some Gaussians parameters in pinned main memory, and loading them to GPU memory on demand. We reduce the overheads from offloading by taking advantage of several unique characteristics of 3DGS.}

\noindent\textbf{3DGS computation is very sparse.} 
3DGS's computation is sparse, in that only a fraction of the scene's Gaussians are used when rendering a view (during training or inference). This is because each view is associated with a camera pose and only those Gaussians within the camera's frustum can contribute to the rendered image, as illustrated by Figure~\ref{fig:frustum-cull}.  In fact, 3DGS's rendering workflow explicitly computes the set of Gaussians within the frustum before processing them for a given view before processing them (shown in Figure~\ref{fig:3dgs-algo} \circled{1}).

\begin{figure}[t]
    \centering
    \includegraphics[width=\columnwidth]{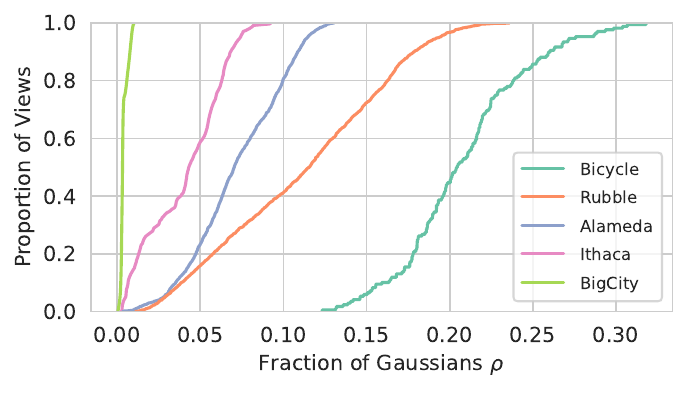}
    \caption{Empirical cumulative distribution functions (CDF) for the sparsity in Bicycle, Rubble, Alameda, Ithaca, and BigCity. 
    }
    \label{fig:sparsity_cdf}
\end{figure}

We have found that a single view accesses a very small fraction (less than 1\%) of a large scene's Gaussians. We quantify this by calculating the sparsity $\mathcal{\rho}^i$ for view $i$ in a scene as $\mathcal{\rho}^i = \frac{|\mathcal{S}_i|}{N}$, where $\mathcal{S}_i$ is the set of Gaussians in view $i$'s and $N$ to be the total number of Gaussians. Figure~\ref{fig:sparsity_cdf} shows the CDF of $\mathcal{\rho}_i$ for the datasets in Table~\ref{tab:scene-aera-resolution-n3dgs-memory}. As can be seen, larger scenes exhibit higher sparsity (aka smaller $\rho$).  
This is expected because, while the number of Gaussians scale as a function of scene size, the volume enclosed by the camera frustum is independent of scene size. 
For the largest scene (Matrixcity BigCity), we found that the average view only accessed $0.39\%$ of Gaussians, and the maximum number of Gaussian's accessed by a view is $1.06\%$.

We leverage sparsity by using 3DGS' frustum culling logic to identify the subset of Gaussians needed to process each view (and thus the microbatch) in advance, and only transfer those needed to the GPU.

\noindent\textbf{Sparsity patterns across views exhibit spatial locality.} 
Different views (for the same scene) have different but overlapping sparsity patterns. In other words, for views $i$ and $j$, $\mathcal{S}_i\neq\mathcal{S}_j$ and the number of  Gaussians in the intersection, $|\mathcal{S}_{i}\cap\mathcal{S}_j|$, is dependent on how much spatial locality exists between these views based on their camera positions and angles.  

We exploit spatial locality to optimize data transfer between CPU and GPU in two ways: 1) we compute each microbatch's sparsity pattern in advance and schedule microbatches carefully to increase overlapped access (\S\ref{sec:design:order-optimization}), and 2) we cache Gaussians used by successive microbatches on the GPU (\S\ref{sec:design:gaussian-caching}), thus reducing communication overheads.

\noindent\textbf{Sparsity patterns can be computed using partial Gaussian information.} 
\label{sec:opportunity:predict-memory-access}
In existing 3DGS implementations, all Gaussian parameters are stored in a single tensor which is used to perform frustum culling on the GPU. 
Doing so requires all Gaussians to be loaded into GPU memory in order to determine a view's sparsity pattern, which contradicts our earlier design choice to only load those necessary Gaussians.
To address this problem, we develop an approach (\S\ref{sec:design:attributewise-offload}) that stores some of the attributes (position, rotation and scale) of all Gaussians on the GPU. As we explain in the next section, this approach is practical because these attributes take relatively little memory.

%% file: 03_design_revision.tex
\section{System Design}\label{sec:design}

\begin{figure}[h!]
    \centering
    \includegraphics[width=1.05\linewidth]{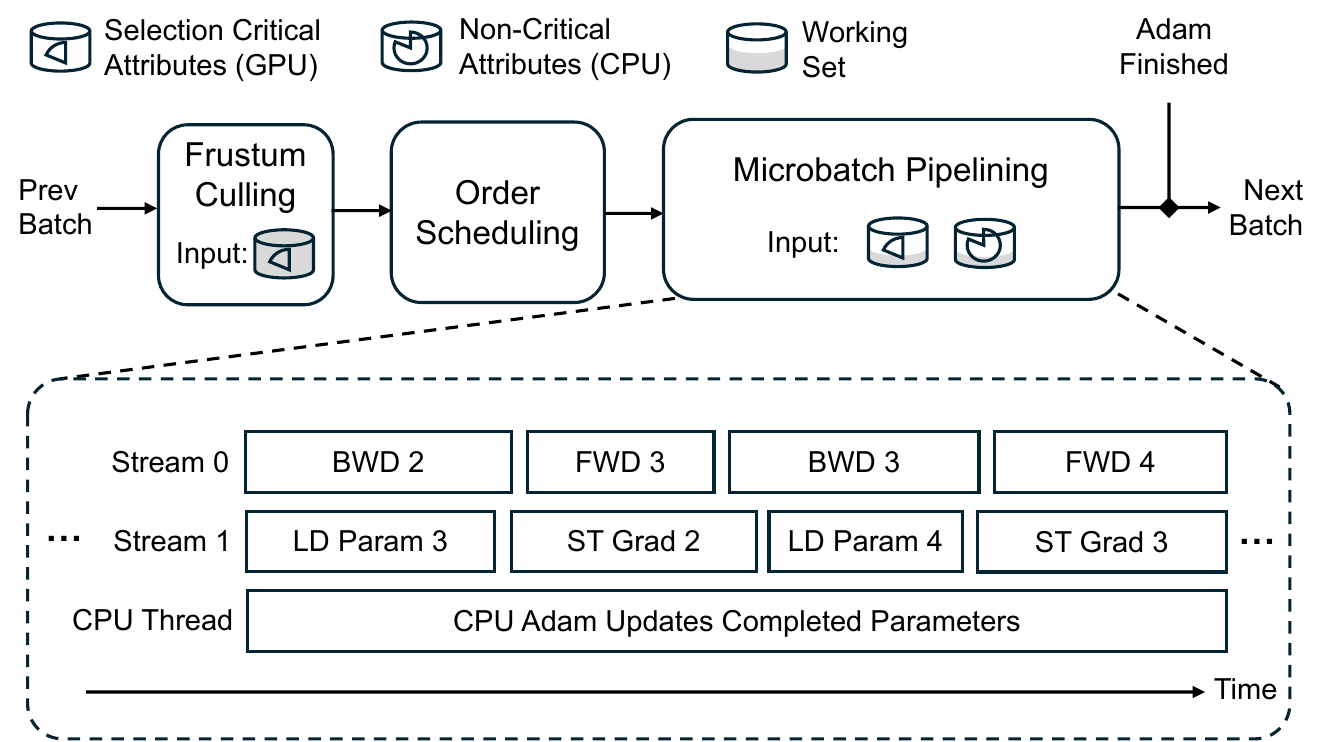} 
    \caption{The workflow of a training step in \name. For a batch of images, \name first performs frustum culling for each image, then schedules their rendering order, and finally uses micro-batch pipelining to overlap communication (on GPU stream 1) and Adam optimizer update (on the CPU) with GPU rendering (on GPU stream 0).
    ``FWD'' and ``BWD'' refer to the forward and backward passes of the i-th microbatch, ``LD'' and ``ST'' refer to loading Gaussian parameters from CPU and storing the gradients to CPU. The numbers (1,2,3) next to these operations indicate the microbatch ID. The area of shading on attributes denotes the proportion that is active as working set.
    }
    \label{fig:system_temporal}
\end{figure}
We now describe \name, a 3DGS system that allows large or highly detailed scenes to be trained/rendered on a single consumer-grade GPU. Our design extends effective GPU memory capacity by offloading Gaussian parameters and optimizer computation to the CPU, while leveraging the observations from the previous section (\S\ref{sec:opportunity:frustum-culling}) to reduce overhead from GPU-CPU communication and CPU computation (\S\ref{sec:naive-offload}). \revise{Our current implementation of \name is built on CUDA, but the design itself is agnostic to the rendering backend and can be ported to the Vulkan platform~\cite{vulkan}.}

\name trains a scene representation as shown in  Figure~\ref{fig:system_temporal}: First, it selects a batch of training images and views, and then uses frustum culling (\S\ref{sec:design:attributewise-offload}) to compute the set of Gaussians $\mathcal{S}_i$ required by each view $i$ in the batch. We refer to these as in-frustum Gaussians. 
Next, it divides a batch into several microbatches to enable microbatch pipelining. More importantly, it uses frustum culling's output to determine the order in which microbatches are processed to maximize spatial locality (\S\ref{sec:design:order-optimization}). Finally, each microbatch is processed in a pipelined fashion to overlap both communication and CPU computation with GPU computation. 
\revise{More concretely, when processing a microbatch, \name loads into GPU memory those in-frustum Gaussians, using Gaussian Caching (\S\ref{sec:design:gaussian-caching}) to avoid redundantly loading any Gaussian that is used by two consecutive microbatches. \name then executes the forward and backward training pass on the GPU and transfers gradients back to the CPU, where a concurrent CPU thread executes the Adam optimizer and updates the Gaussian parameters. As Figure~\ref{fig:system_temporal} shows, the CPU-GPU communication for loading in-frustum Gaussians for microbatch $i$ overlaps with the backward GPU computation for microbatch $i-1$; and the GPU-CPU communication for transferring gradients of microbatch $i$ overlaps with the forward GPU computation for microbatch $i+1$. 
For those Gaussians that are last updated by a microbatch, \name performs their corresponding Adam updates on CPU, which overlaps with the forward/backward GPU computation done by subsequent microbatches (\S\ref{sec:design:overlap-adam}).}

\subsection{Attribute-Wise Offload} \label{sec:design:attributewise-offload} 
As we discussed previously in \S\ref{sec:opportunity:frustum-culling}, the frustum culling step is run on the GPU and requires access to some attributes (e.g., position) of all Gaussians in the scene. As our goal is to scale to scenes whose Gaussians (by which we mean all attributes) do not fit in GPU memory, we cannot load all Gaussians into GPU memory before running the frustum culling step.

We address this by observing that frustum culling accesses a small subset of each Gaussian's attributes: For each Gaussian, the frustum culling algorithm checks the intersection between the view frustum and the Gaussian. When computing intersection, the algorithm only needs information on the Gaussian's position, scale, and rotation. This is because intersection does not depend on the Gaussian's color (determined by the spherical harmonics) or opacity. 
\revise{In particular, during frustum culling, 3DGS determines whether a Gaussian is in-frustum by computing the intersection between the view frustum and the Gaussian's ellipsoid (which is derived from its scale and rotation) and checking whether it is within $3$ standard deviations ($3\sigma$) of the Gaussian's distribution. Culling within $3\sigma$ is standard practice in existing 3DGS implementations~\cite{3dgs, gsplat}.} 
We refer to the attributes required for frustum culling as \emph{selection-critical attributes}, and the rest of the attributes as \emph{non-critical attributes}.

We observe that the selection-critical attributes constitute less than 20\% (10 out of 59 floats) of a Gaussian's memory footprint (Table~\ref{tab:3dgs_attributes}). Thus, \name keeps the selection-critical attributes for all Gaussians in GPU memory, i.e., they are never swapped out to CPU memory, and no additional CPU-GPU communication is necessary for frustum culling. Non-critical attributes are stored in CPU memory, and loaded into GPU memory only when required.

\subsection{Microbatch Pipelining} \label{sec:design:microbatches-pipeline} 
3DGS training uses minibatch gradient descent. In contrast to existing 3DGS systems that process a whole batch at a time, our design divides each batch into several minibatches to use pipelining and gradient accumulation~\cite{gpipe, pipedream}. 
In our setting, a batch consists of multiple images and a microbatch consists of a single image. We start the forward pass for microbatch $i+1$ as soon as the backward pass for microbatch $i$ completes.

\name's use of microbatch pipelining reduces GPU memory requirements: each forward and backward pass processes a single image at a time, reducing the amount of activation memory required. More importantly, as shown in Figure~\ref{fig:system_temporal}, microbatch pipelining allows \name to overlap communication for one microbatch with the computation for another, thereby hiding communication overhead. \name uses double-buffering to ensure that communication and computation can be safely overlapped. While the use of double-buffering increases memory requirements, the additional memory requirements are independent of scene and batch size.

We further improve basic microbatch pipelining by incorporating three domain specific optimizations, which are described next:
\begin{inparaenum}[(a)]
    \item Precise Gaussian Caching (\S\ref{sec:design:gaussian-caching}); 
    \item Overlapped CPU Adam (\S\ref{sec:design:overlap-adam}); and
    \item Pipeline Order Optimization (\S\ref{sec:design:order-optimization}). 
\end{inparaenum}

\subsubsection{Precise Gaussian Caching} \label{sec:design:gaussian-caching} 

Our first optimization builds on the observation that some of the Gaussians accessed by microbatch $i$ and $i+1$ are the same because of spatial locality. Since the frustum culling step has already computed in-frustum Gaussians for each microbatch, \name uses this information to reduce the number of Gaussians loaded into GPU memory from CPU memory: when loading Gaussians for microbatch $i+1$, it copies the intersecting Gaussians (aka those in $\mathcal{S}_i\cap\mathcal{S}_{i+1}$) from microbatch $i$'s Gaussian parameter tensors (which are already in GPU memory), and only loads those Gaussians not in the intersection from CPU memory. Note that copying Gaussians in this way does not require additional memory beyond what is already allocated for double buffering. 

We also use the same approach to avoid redundant copies when transferring gradients from GPU memory to CPU memory: after microbatch $i$ has finished, we only transfer the gradients that are not going to be updated by microbatch $i+1$, i.e., gradients for Gaussians in the set $\mathcal{S}_{i}\setminus \mathcal{S}_i\cap\mathcal{S}_{i+1}$. We copy the rest of the gradients for Gaussians $\mathcal{S}_i\cap\mathcal{S}_{i+1}$ into microbatch $i+1$'s gradient buffer to be accumulated.

\subsubsection{Overlapped CPU Adam}\label{sec:design:overlap-adam}
\begin{figure}[t]
    \centering
    \includegraphics[width=1\linewidth]{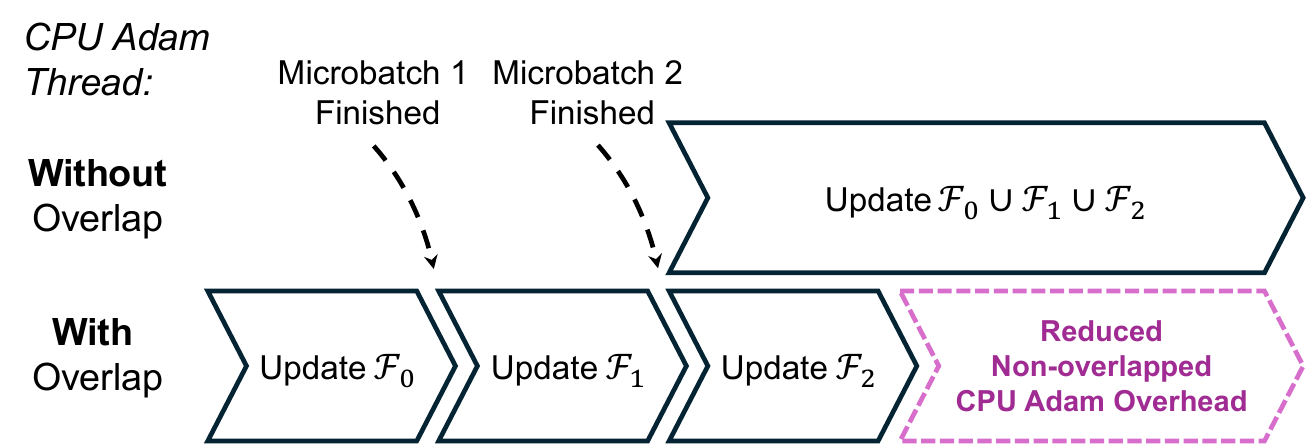}
    \caption{Illustration of overlapping CPU Adam with batch size 2. The upper half of this image illustrates the overhead caused by CPU Adam, while the lower half demonstrates the benefits of overlapping CPU Adam. Within the diagram, $\mathcal{F}_i$ represents the set of Gaussians that have just finalized before microbatch $i$ and are ready to update their parameters using Adam. For example, $\mathcal{F}_0$ includes Gaussians that are not affected by any images in the batch. $\mathcal{F}_{|Batch|}$ consists of the Gaussians touched in the last microbatch, which cannot be overlapped.}
    \label{fig:overlap-cpu-adam}
\end{figure}
At the end of each batch, 3DGS training uses the Adam optimzer~\cite{kingma2015adam} to combine the computed gradients and update Gaussian parameters. We observe that many Gaussian updates can be \emph{finalized} early before the last microbatch. In other words, in a batch of size $B$, the last microbatch ($i$) that accesses (and thus updates) some Gaussian $g$ might be $i<B$. In this case, it is safe to use Adam to update Gaussian $g$ before the whole batch completes. Doing such early update is desirable because \name performs Adam update on the CPU, which can be overlapped with the forward and backward GPU computation of subsequent microbatches. \name implements this optimization (Figure~\ref{fig:overlap-cpu-adam}): When scheduling a batch, for each Gaussian $g$, it computes the microbatch $L_g$ at which $g$ is finalized by computing $L_g = \max \{i \mid g \in \mathcal{S}_{i} \}$ ($L_g=0$ if $g$ is not accessed by a batch). At the end of every microbatch $j$, \name uses CPU Adam to update all Gaussians $g$ whose $L_g = j$. Only those Gaussians that are finalized in the last microbatch ($L_g=B$) do not have their CPU Adam computation overlapped.

\subsubsection{Pipeline Order Optimization} \label{sec:design:order-optimization} 
The order in which microbatches are processed within a batch does not affect correctness. \revise{This is because gradients calculated for individual microbatches are accumulated over the full batch before applying optimizer update, and thus the same final update is computed for a batch regardless of its microbatch ordering.}
However, the order of microbatches determines the effectiveness of Gaussian caching and overlapped CPU Adam.  If the microbatch schedule leads to a large overlap between Gaussians accessed by consecutive microbatches (i.e., when $\mathcal{S}_{i+1}\cap\mathcal{S}_{i}$ is large for all $i$), Gaussian caching can eliminate more communication.  Similarly, if the schedule results in a larger number of Gaussians finalized in earlier microbatches,  more of the CPU Adam computation can be overlapped. Therefore, \name tries to find a microbatch schedule that maximizes the effectiveness of both optimizations. 

Scheduling of microbatch computation is aided by the observation that 3DGS exhibits spatial locality (\S\ref{sec:opportunity:frustum-culling}). As such, any schedule that maximizes Gaussian overlap between consecutive microbatches must process all views in the same region in close temporal proximity. Further, Gaussians are finalized once all views in a region have been rendered. Thus, schedules that maximize overlap also tend to finalize a large number of Gaussians early. Given this observation, we compute a good schedule by formulating the scheduling problem as an instance of the Traveling Salesman Problem (TSP)~\cite{tsp}: we treat each microbatch as a node in the graph, and the distance between two microbatches $i$ and $j$ is given by the symmetric difference between the Gaussians accessed by each ($|\mathcal{S}_{i} \oplus \mathcal{S}_{j}|$ which gives the number of Gaussians that do not overlap). TSP computes the shortest Hamiltonian path through this graph, which by construction is the schedule that maximizes overlap. Our implementation uses stochastic local search with a greedy heuristic~\cite{2opt} to quickly generate an optimal solution; see Appendix~\ref{sec:appendix-tsp} for details about our formulation and the greedy search algorithm.

\section{Implementation} \label{sec:impl}

We implemented \name by extending Grendel~\cite{grendel}, an existing 3DGS training framework. Our extensions added the offloading approach described in the previous section and incorporated the rasterization kernels of gsplat~\cite{gsplat} into Grendel. We discuss important details about the implementation below.

\subsection{Pre-rendering Frustum Culling}
\label{sec:implement:frustum-cull}

In current implementations of 3DGS\cite{3dgs}\cite{gsplat}, the frustum culling step is fused to the rendering kernels. These rendering kernels process all Gaussians as input, but only utilize those that intersect with the Frustum. The intersected Gaussians are computed implicitly by cuda threads and registers, without being explicitly stored in GPU memory. In large scenes with low $\rho$, the majority of input Gaussians are not in-frustum and hence result in substantial wasted GPU computation. Storing intermediate activations for non-in-frustum Gaussians also wastes GPU memory. The backward pass also performs unnecessary computation because it calculates gradients for the entire Gaussian input tensor, even though only in-frustum Gaussians have non-zero gradients. 

In our implementation, we perform frustum culling to obtain in-frustum Gaussians indices $\mathcal{S}_i$ and store them in GPU memory before rendering. This allows us to explicitly eliminate unnecessary Gaussians from the input to the rasterization kernels, thus decreasing the input size by $\rho_i$, reducing $N$ to $|\mathcal{S}_i|$. Doing so reduces both GPU memory and computation usage.

Pre-rendering frustum culling is a simple engineering method that can also be applied to traditional GPU-only training without offloading. The evaluations in Section \ref{sec:eval:speed} demonstrate that this engineering technique reduces memory usage and significantly enhances throughput when training a highly sparse scene. We will also show that \name remains highly effective even after eliminating the influence of this engineering trick.

\subsection{Selective Loading Kernel} \label{sec:selective-loading-kernel} 

After computing indices for the in-frustum Gaussians, we use a custom kernel to load parameters from CPU memory. As in-frustum Gaussians are spread over CPU memory because of the sparse access pattern, naively copying them individually using \texttt{cudaMemcpy} (or \texttt{cudaMemcpyAsync}) underutilizes CPU-GPU communication bandwidth. Instead, our implementation stores offloaded Gaussian attributes in pinned CPU memory that can be accessed directly from CUDA code running on the GPU without and we develop a \emph{selective loading kernel} which loads (over PCIe) the in-frustum Gaussian parameters from CPU memory to GPU registers and then stores the register values into GPU memory. Since all of the communication is initiated from the GPU, this kernel avoids CPU-GPU coordination. In addition, the same kernel is also used to copy cached Gaussians from GPU memory for processing.

To further improve communication efficiency, we concatenate and pad attribute tensors when storing them in CPU memory so that all attributes of a Gaussian are stored in contiguous memory and are cache-line aligned. The selective loading kernel splits Gaussians attributes when loading them into GPU memory and concatenates those in-frustum Gaussian attributes together. Implementing splitting and concatenation logic in the same kernel reduces computational overheads.

We also develop a similar kernel to efficiently transfer gradients from the GPU to CPU memory.

\subsection{Separate Communication Stream} \label{sec:implement:comm-stream}

For pipelined execution of microbatches, we employ two CUDA streams: one for computation, and the other for communication. As illustrated in Figure \ref{fig:system_temporal}, the parameters loading and gradients storing are interleaved in the communication stream, analogously to the 1F1B pipelining method for training neural networks described in \cite{1f1b}. We add CUDA events to correctly synchronize operations across streams, making sure that in-frustum gaussians parameters are loaded before executing the microbatch and gradients are offloaded after finishing the backward pass. We increase the communication stream priority over the computation stream to prevent delays in executing the communication kernel by GPU. These delays, as we observed, impede subsequent microbatch processing and ultimately slow down the overall training process. 

The kernel responsible for gradient offloading needs a minor adjustment to support gradient accumulation in pipelining. In the gradient offloading kernel, rather than directly storing gradients' values to CPU memory, it first fetches the previously accumulated gradients, adds them within cuda registers, and then stores the sum back in CPU memory. %

Our pipeline approach involves prefetching parameters for the upcoming microbatch while postponing gradient offloading from the prior one. We use double buffer to achieve this, which can raise memory usage. However, with our single communication stream and 1F1B setup, along with precise timing in managing creation and deallocation, we prevent the coexistence of double buffers for previous microbatch gradients and upcoming microbatch parameters. This effectively controls the additional memory usage from double buffer.

\subsection{Thread for CPU Adam update} \label{sec:implement:cpu-adam}

Our CPU Adam implementation extends the Zero-offload \cite{zerooffload} implementation to allow updating a subset of Gaussians (which have completed gradients at each call). We execute CPU Adam on a dedicated thread CPU thread. To allow concurrent execution, we release the Python GIL lock within the CPU adam thread, allowing the primary Python thread to continue processing. A signal buffer in CPU pinned memory is used to synchronize the GPU communication stream and the CPU Adam thread. Specifically, the GPU communication stream sets the gradient completion signal via DMA after the gradient transfer kernel finishes, while the CPU Adam thread waits on the signal buffer before performing the Adam update after the microbatch. %

%% file: 04_eval_revision.tex
\section{Evaluation} \label{sec:eval}

In this section, we evaluate \name and the following are the highlights of our results: 
\begin{itemize}
    \item \name enables 3DGS training of models up to 6.1x larger through CPU offloading, compared to GPU-only training baselines. 
    \item \name enhances reconstruction quality by training larger model, achieving PSNR 25.15 for BigCity\cite{matrixcity} using 102 million Gaussians. In contrast, the GPU-only training reaches a PSNR of 23.93, using only 15 million Gaussians to avoid running out of memory.
    \item \revise{
    \name has modest offloading overhead. It achieves 86\%--97\% of the throughput of an enhanced GPU-only baseline on RTX 2080 Ti and 55\%--90\% on  RTX 4090. Compared to naive offloading, \name is 1.38 to 1.92 faster.}

\end{itemize}

\subsection{Setting and Datasets}

\noindent\textbf{Testbeds.}
\revise{We run our evaluation on two testbeds: One is a machine with an AMD Ryzen Threadripper PRO 5955WX 16-core CPU, 128 GB RAM, and a 24 GB NVIDIA RTX 4090 GPU connected over PCIe 4.0. The other is a machine with Intel Xeon E5-2660 v3 20-Core CPU, 256 GB RAM, and a 11 GB NVIDIA RTX 2080 Ti GPU connected over PCIe 3.0.}

\revise{These two settings allow us to evaluate \name under different computation and communication speed. In particular, the RTX 2080 Ti has about 7x fewer FLOPs (cuda core) than the RTX 4090, and PCIe 3.0 has 2x less bandwidth than PCIe 4.0. Therefore, in our experiments, the 2080 Ti GPU testbed is more likely to be compute bound.
}

\noindent\textbf{Datasets.} 
Our evaluation uses the datasets presented in Table~\ref{tab:scenes-main}. These datasets cover a variety of scene sizes (which roughly correlate with the number of images), image resolution, and scene types that collectively represent diverse workloads. 
Specifically, the scene size affects both the Gaussian model size and the degree of sparsity $\rho$ in rendering; resolution affects both GPU rendering speed and activation memory usage; additionally, the scene topology determines the sparsity patterns. Together, they provide a comprehensive evaluation of our system. %
We detail the preparation of the datasets in the Appendix \ref{appendix:datasets-prepare}.

\begin{table}[h]
\centering
 \resizebox{\columnwidth}{!}{
\begin{tabular}{lrrrrr}
    \toprule
    \textbf{Scene} & \textbf{\# Images} & \textbf{Resolution} & \textbf{Scene Type} & \textbf{BS}\\
    \midrule
    Bicycle~\cite{mipnerf360}& 200 & 4K& Yard & 4 \\
    Rubble~\cite{meganerf} & 1600 & 4K & Aerial & 8 \\
    Alameda~\cite{zipnerf} & 1700 & 2K  & Indoor & 8 \\
    Ithaca365~\cite{ithaca365} & 8200 & 1K & Street  & 16 \\
    MatrixCity BigCity~\cite{matrixcity} & 60000 & 1080P & Aerial & 64 \\
    \bottomrule
\end{tabular}
}
\caption{Scenes used in our evaluation: Our selection includes scenes of different sizes, resolutions, and types, representing a diverse range of workload characteristics. We set the training batch size (BS) in our experiments to correspond with their scene sizes. } 
\label{tab:scenes-main}
\end{table}

\noindent\textbf{Baseline.} 
We choose Grendel-GS \cite{grendel} as our baseline. Although Grendel-GS is a multi-GPU training system, we run in its single-GPU mode to take advantage of its efficient training framework. We use GSplat's CUDA kernels\footnote{These CUDA kernels consume over 95\% time in 3DGS training.}~\cite{gsplat} as the rendering backbone for its memory efficiency by setting the corresponding flag in the Grendel-GS training framework. We refer to this GPU-only implementation as ``baseline''. %

\revise{
\noindent\textbf{Enhanced Baseline.}
Additionally, we build an enhanced version of the baseline by adopting \name's pre-rendering frustum culling feature (\S\ref{sec:implement:frustum-cull}) to avoid unnecessary kernel computation. This version is closer to \name's kernel implementation, and thus provides a better baseline for evaluating \name's offloading overhead. We refer to this GPU-only variant as ``enhanced baseline''.
}

\noindent\textbf{Naive Offloading.}
We implement the naive offloading strategy (see \S\ref{sec:naive-offload} and Figure~\ref{fig:offload-simple}) on Grendel \cite{grendel}. The implementation uses pinned memory for GPU-CPU communication as in \name. It also utilizes the same CPU Adam as \name (\S\ref{sec:implement:cpu-adam}) and adopts \name's pre-rendering frustum culling technique for efficient kernel computation. Additionally, it trains each batch one image at a time with gradient accumulation to reduce activation memory usage. By comparing with this naive offloading, we can quantify how \name's various offloading techniques can improve performance. 

\subsection{Memory Efficiency} \label{sec:eval:memory} 

\begin{figure*}[t]
    \centering
    \begin{subfigure}[t]{0.49\textwidth}
        \includegraphics[width=\linewidth]{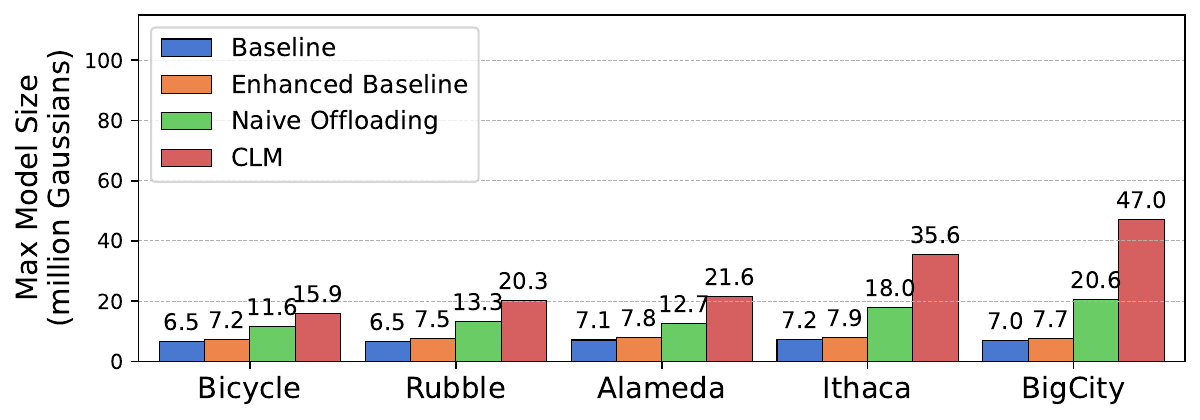}
        \caption{RTX 2080 Ti}
        \label{fig:eval-max-model-size_2080}
    \end{subfigure}
    \hfill
    \begin{subfigure}[t]{0.49\textwidth}
        \includegraphics[width=\linewidth]{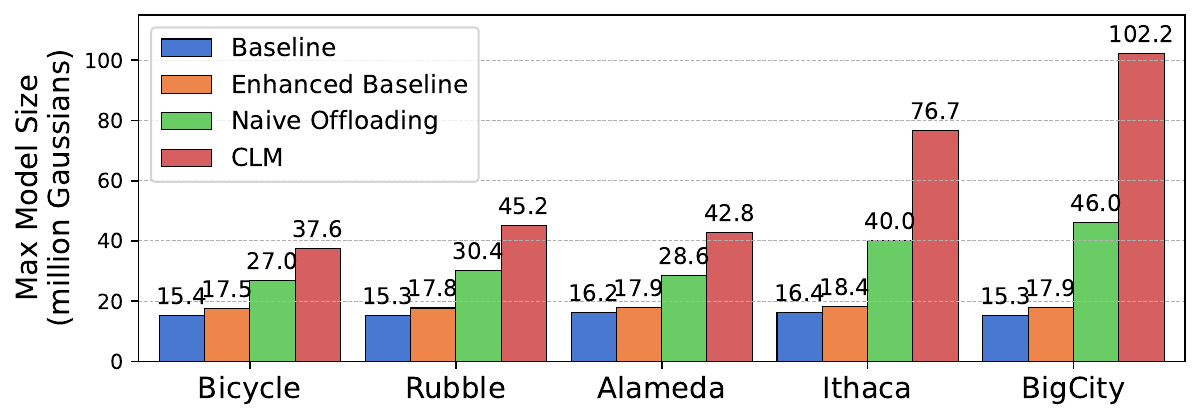}
        \caption{RTX 4090}
        \label{fig:eval-max-model-size_4090}
    \end{subfigure}
    
    \caption{\revise{Maximum trainable model size without OOM across two GPU testbeds and five scenes. \name consistently enables significantly larger models. The BigCity scene shows the most notable improvement---6.1x larger than the enhanced baseline and 2.3x over naive offloading on RTX 2080 Ti; and 5.7x and 2.2x larger respectively on RTX 4090.
    } } 
    \label{fig:eval-max-model-size}
\end{figure*}

\name is able to push the model size trainable on a single GPU by up to 6.1 times compared to the enhanced baseline\footnote{We enable the PyTorch's \texttt{expandable\_segments} feature in all experiments to minimize the impact of GPU memory fragmentation.}.  Larger models enhance the quality of reconstruction, achieving state-of-the-art PSNR for the BigCity scene with 102 million Gaussians. 

\noindent\textbf{Larger model training made possible. }
\revise{
Figure~\ref{fig:eval-max-model-size} shows the maximum model size that could be trained before encountering an OOM error on each testbed.  We can see that \name allows 
larger model sizes to be trained across all scenes. 
Specifically, the GPU-only baseline can support a maximum of 7.2 M and 16.4 M Gaussians on RTX 2080 Ti and RTX 4090, respectively, before running out of memory. 
The enhanced baseline avoids storing the activations of Gaussians unused in rendering via pre-rendering frustum culling (\S\ref{sec:implement:frustum-cull}), thereby postponing the OOM point to 7.9 M and 18.4 M Gaussians, respectively. 
Using CPU memory, naive offloading can support up to 20.6 M and 46 M Gaussians before exhausting GPU memory. 
In contrast, \name supports up to 47 M and 102.2 M Gaussians---up to 6.1x larger than the enhanced GPU-only baseline, and 2.3x larger than naive offloading. }

\revise{
\name requires less memory than naive offloading because it does not load all Gaussian parameters to GPU memory before each rendering step.
The difference in the maximum supported model sizes by \name on 2080 Ti vs 4090 (e.g., 47 M vs. 102.2 M Gaussians for BigCity) roughly reflect their GPU memory capacities (11 GB vs. 24 GB). 
We also observe that the maximum trainable model size is dependent on the scene. In particular, scenes that have higher resolution (see Table \ref{tab:scene-aera-resolution-n3dgs-memory}) or worse sparsity (i.e., high $\rho$, as shown in Figure \ref{fig:sparsity_cdf})---such as Bicycle, Rubble---require more activation memory, leaving less memory available for Gaussian parameters compared to scenes like Ithaca and BigCity, which have lower resolution and lower $\rho$. 
}

\begin{figure}[t]
    \centering
    \includegraphics[width=0.8\linewidth]{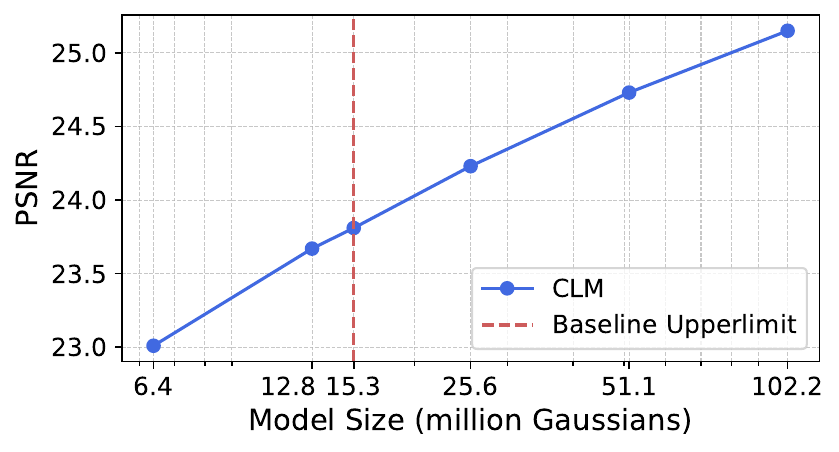}
    \caption{Scalability on BigCity. X axis (log scale) is the model size presented in million Gaussians. Y axis (linear) is the PSNR evaluated on rendered images. Each data point is trained using \name on a single 24GB RTX 4090 GPU for 500000 steps. This shows \name enables scaling to 102 million Gaussians and achieve a state-of-the-art PSNR of 25.15. }
    \label{fig:scalability_BigCity}
\end{figure}

\noindent\textbf{Larger models improve reconstruction quality.}
This experiment assesses the importance of scalability model sizes using the BigCity scene\cite{matrixcity}, a city-scale benchmark covering 25.3 ${km^2}$ with extensive details.
We train models consisting of 6.4 M, 12.8 M, 15.3 M, 25.6 M, 51.1 M, and 102.2 M Gaussians, doubling the size incrementally. Among these, the 15.3 M model is the largest one that can be trained by the GPU-only baseline on the 24 GB RTX 4090 testbed before running out of memory. We train each of these models for 500,000 steps using \name and evaluate the reconstruction quality using peak signal-to-noise ratio (PSNR), where a higher PSNR indicates better reconstruction quality (aka the rendered image is closer to the ground truth). 
Figure~\ref{fig:scalability_BigCity} shows that \name improves reconstruction quality by allowing larger models to be trained using additional CPU memory. With a model size of 102.2 M Gaussians, \name achieves a PSNR of 25.15. In contrast, the GPU-only baseline is limited to 15.3 M Gaussians and yields a lower PSNR of 23.93.

\noindent\textbf{Breakdown of GPU memory consumption.}
We examine different systems' GPU memory consumption statistics when training the Rubble and BigCity scenes. We use model sizes 15.3 M, 30.4 M, and 45.2 M Gaussians for Rubble; and 15.3 M, 46.0 M, and 102.2 M for BigCity, corresponding to the maximum supported model sizes of the baseline, naive offloading and \name, respectively (see Figure~\ref{fig:eval-max-model-size}).

\begin{figure}[t]
    \centering
    \begin{subfigure}[t]{\linewidth}
        \centering
        \includegraphics[width=1\linewidth]{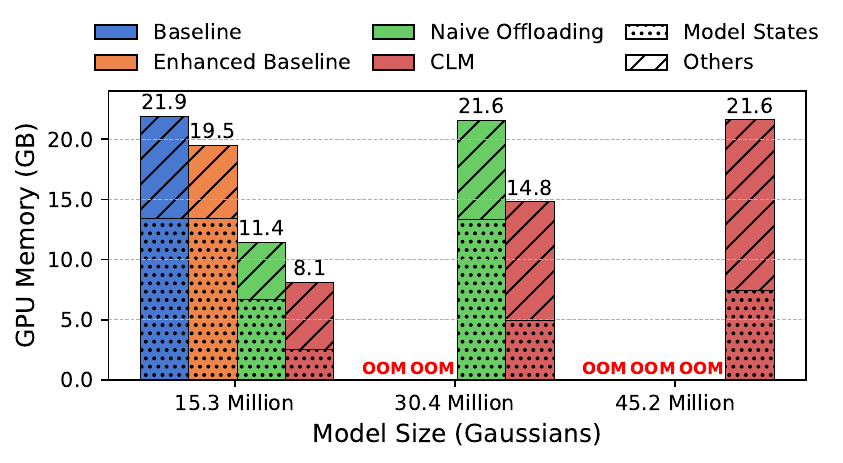}
        \caption{Rubble}
        \label{fig:rubble-memory-decomposition}
    \end{subfigure}
    \hfill
    \begin{subfigure}[t]{\linewidth}
        \centering
        \includegraphics[width=1\linewidth]{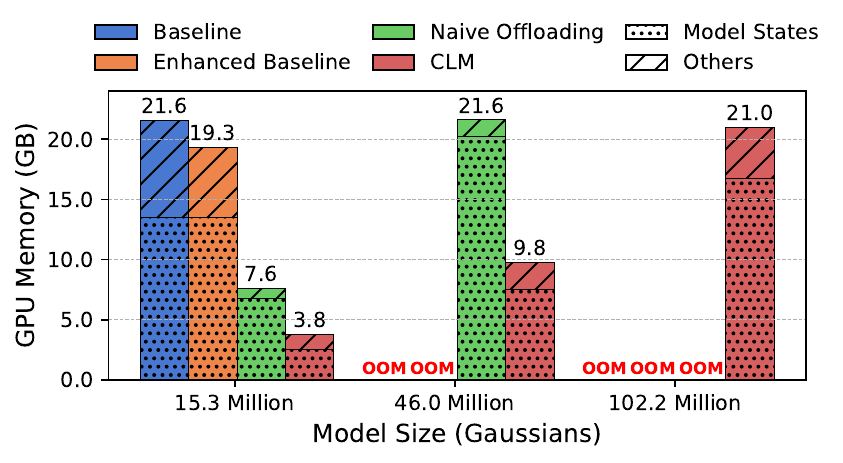}
        \caption{BigCity}
        \label{fig:bigcity-memory-decomposition}
    \end{subfigure}
    
    \caption{Memory usage breakdowns for Rubble using varying model sizes on the 4090 testbed. Each bar is composed of two parts: model states (bottom) and others (upper). \name consumes the least GPU memory, thus avoiding OOM.}
    \label{fig:memory-decomposition}
\end{figure}

For the Rubble scene (see Figure~\ref{fig:rubble-memory-decomposition}), during the training of the 15.3 M model  where all four systems operate without running out of memory, the baseline consumes the most GPU memory, whereas \name requires the least. The enhanced baseline and naive offloading fall somewhere in the middle. 
The two baselines consume the same amount of model state memory, with a small difference in "others" because the pre-rendering frustum culling (\S\ref{sec:implement:frustum-cull}) in the enhanced baseline removes redundant activation states.
For the 30.4 M and 45.2 M models, the baselines encounter GPU OOM. In contrast, both naive offloading and \name support the training of the 30.4 M model, illustrating the advantages of GPU memory saving through offloading. 
Despite keeping some selection-critical attributes of all Gaussians on the GPU, \name uses less GPU memory than naive offloading, which offloads all attributes to CPU memory. This is because naive offloading transfers all attributes of all Gaussians to the GPU before each rendering, which consumes larger GPU memory. 
While \name significantly cuts down memory for Gaussian model states, it slightly elevates the ``others'' memory usage. This is due to the double buffer design in our pipelining (\S\ref{sec:implement:comm-stream}) which enables parameters prefetching and delayed gradient offloading. 
All of the above observations can be drawn similarly for the BigCity scene in Figure \ref{fig:bigcity-memory-decomposition}. The primary difference is that BigCity has a smaller $\rho$ and lower resolution compared to Rubble, leading to model states memory dominating activation memory. Consequently, offloading model state in BigCity results in a more substantial overall memory reduction.

\subsection{Performance} \label{sec:eval:speed} 

We compare \name's training performance to that of naive offloading to evaluate the effectiveness of \name's design choices. We also quantify \name's offloading overhead by comparing its performance to GPU-only baselines.
We measure performance by training throughput,  calculated as the number of images processed per second during training. 

\begin{figure}[t]
    \centering
    \begin{subfigure}[t]{\linewidth}
        \centering
        \includegraphics[width=\linewidth]{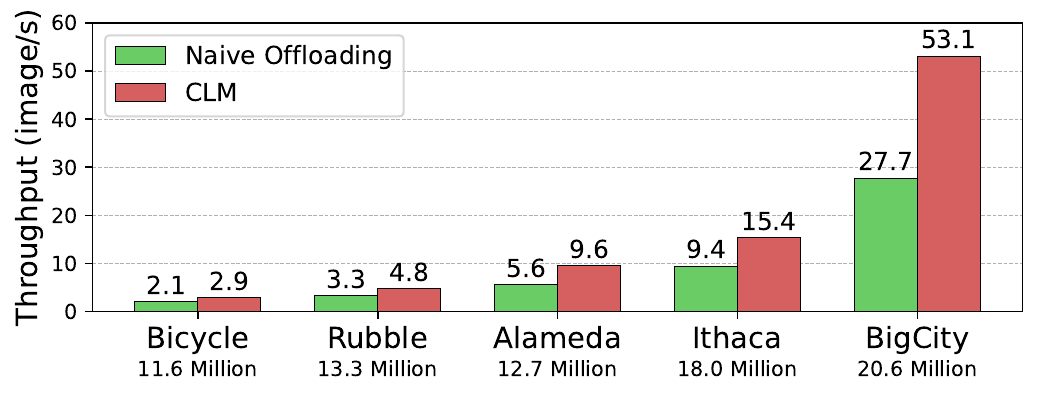}
        \caption{RTX 2080 Ti}
        \label{fig:eval-throughput-compare-naive-offloading_2080}
    \end{subfigure}
    \hfill
    \begin{subfigure}[t]{\linewidth}
        \centering
        \includegraphics[width=\linewidth]{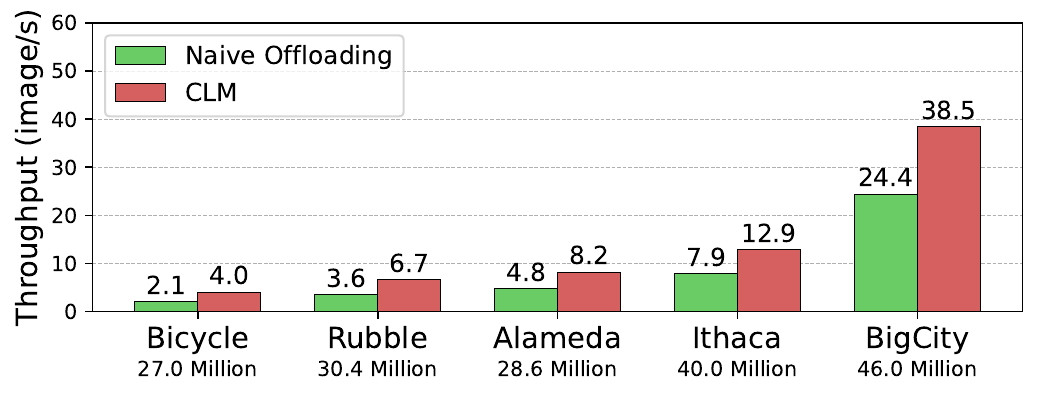}
        \caption{RTX 4090}
        \label{fig:eval-throughput-compare-naive-offloading_4090}
    \end{subfigure}

    \caption{
    \revise{The training throughput of \name vs. naive offloading.
    For each scene-testbed pair, we use the largest model size supported by naive offloading from Figure \ref{fig:eval-max-model-size} to avoid OOM. 
    \name achieves up to 1.92x (BigCity) and 1.90x (Bicycle) speedup on RTX 2080 Ti and RTX 4090, respectively. 
    } } %
    \label{fig:eval-throughput-compare-naive-offloading}
\end{figure}

\noindent\textbf{\name vs. naive offloading.}
We evaluate CLM's performance compared to naive offloading \revise{on both testbeds. Our evaluation covers all scenes in Table~\ref{tab:scenes-main}. For each scene, we use the largest model size that could be trained using naive offloading on the given testbed (Figure~\ref{fig:eval-max-model-size}). As can be seen in Figure~\ref{fig:eval-throughput-compare-naive-offloading}, CLM achieves significant speedup over naive offloading.  In particular, for the largest scene BigCity, CLM is 1.92x faster than naive offloading when run on the 2080 Ti, and 1.58x faster on the 4090.
As can been observed, \name's speedups differ between the two testbeds. As RTX 2080 Ti has roughly 7 times fewer FLOPS than the 4090, the GPU computation takes longer on the 2080 Ti, allowing \name to overlap and hide more of the offloading overhead than on the RTX 4090. This effect is particularly observable in large scenes such as BigCity with high offloading overheads, thus \name has larger improvements when run on a slow GPU than on a faster GPU. %
}

\noindent\textbf{\name vs. GPU-only training.}
We evaluate the performance of \name compared to GPU-only baselines. \revise{To avoid OOM, for each scene in Table~\ref{tab:scenes-main}, we use the maximum model size that can be trained using the baseline (Figure~\ref{fig:eval-max-model-size}).} 

\begin{figure}[t]
    \centering
    \begin{subfigure}[t]{\linewidth}
        \centering
        \includegraphics[width=\linewidth]{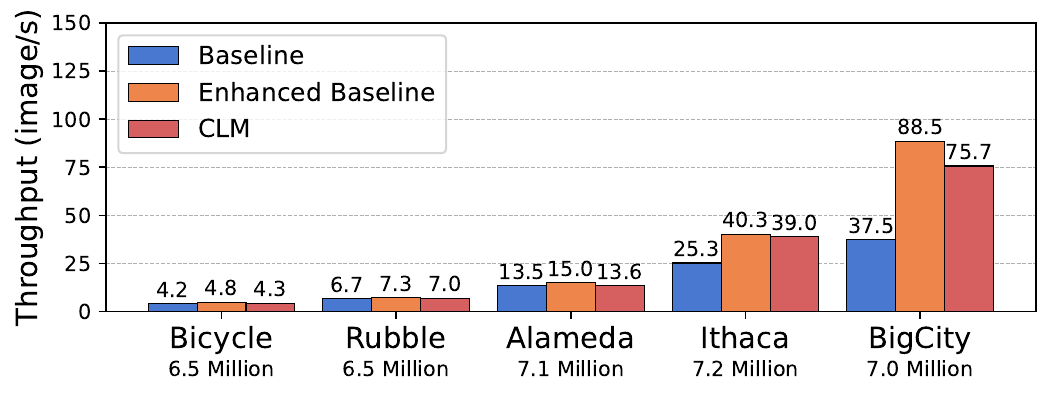}
        \caption{RTX 2080 Ti}
        \label{fig:eval-throughput-small_2080}
    \end{subfigure}
    \hfill
    \begin{subfigure}[t]{\linewidth}
        \centering
        \includegraphics[width=\linewidth]{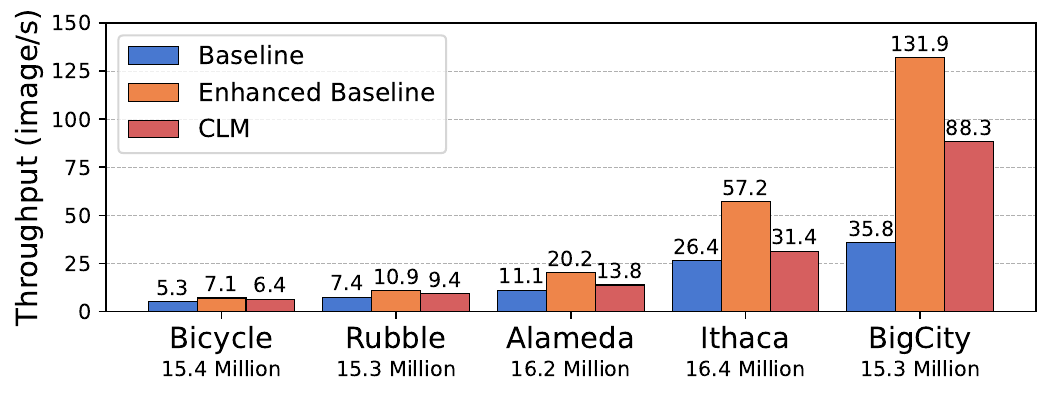}
        \caption{RTX 4090}
        \label{fig:eval-throughput-small_4090}
    \end{subfigure}
    
    \caption{\revise{The training throughput of \name training vs. GPU-only baselines. For each scene-testbed pair, we use the largest model size supported by the baselines as assessed in Figure \ref{fig:eval-max-model-size} to prevent OOM.  Comparing \name to enhanced baseline eliminates the effect of pre-rendering frustum culling.
    On RTX 2080 Ti, \name reaches 86\% (BigCity) to 97\% (Ithaca) of the enhanced baseline's throughput; on RTX 4090, it achieves 55\% (Ithaca) to 90\% (Bicycle). 
    } 
    }
    \label{fig:eval-throughput-compare-baseline}
\end{figure}

\revise{As shown in Figure~\ref{fig:eval-throughput-compare-baseline}, \name achieves similar or much better throughput than the naive baseline on both testbeds.}
The unexpected improvement compared to the baseline is due to \name's use of pre-rendering frustum culling as explained in \S\ref{sec:implement:frustum-cull}. 
This technique enables more efficient computation in scenes with low $\rho$ (e.g., BigCity) by culling more points, resulting in notable performance improvements. 

\revise{We can evaluate \name's offloading overheads more fairly by comparing to the enhanced baseline, which also uses (and benefits from) pre-rendering frustum culling.
As can be seen in Figure \ref{fig:eval-throughput-compare-baseline}, \name achieves 86\% (BigCity) to 97\% (Ithaca) of the enhanced baseline's throughput on the RTX 2080 Ti, and 55\% (Ithaca) to 90\% (Bicycle) on the RTX 4090.
The slowdown occurs for \name because the communication and CPU Adam computation cannot be fully overlapped with GPU computation.
As expected, the offloading overheads depend on both testbed and scene characteristics.
Among the two testbeds, we observe larger overheads on the RTX 4090 than the RTX 2080 Ti because the longer GPU computation time on the 2080 allows \name to more effectively overlap communication and CPU Adam computation.
In terms of scenes, we find that scenes that require more GPU computation (e.g., because they have higher resolution) or where less communication is required (e.g., because of lower $\rho$) have smaller overheads, because communication and CPU Adam computation can be more effectively overlapped. For example, the Rubble and Bicycle scenes which are at 4K resolution have a slowdown of less than 20\% compared to the enhanced baseline, while Ithaca which has the lowest resolution among the scenes we evaluated  (Table~\ref{tab:scene-aera-resolution-n3dgs-memory}) has a slowdown of 45\%.
}

\begin{figure}[t]
    \centering
    \includegraphics[width=\linewidth]{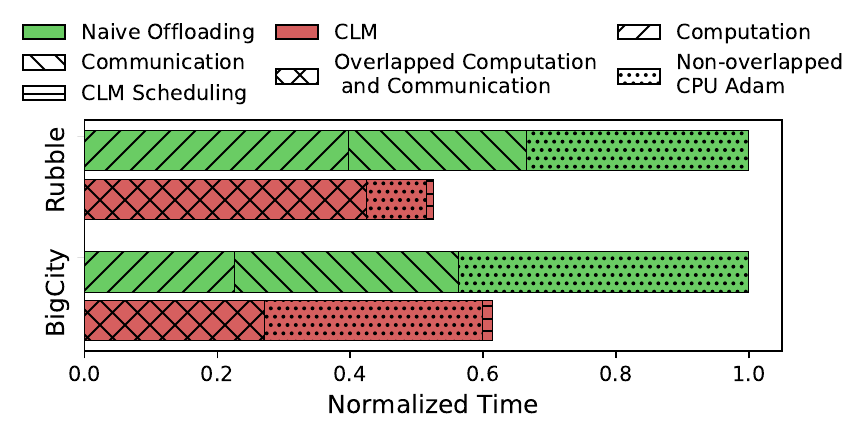}
    \caption{Runtime Decomposition for Rubble and BigCity using \name and naive offloading on the 4090 testbed. We normalize using the total time of naive offloading in each scene. }
    \label{fig:eval_perf_decomp}
\end{figure}

\noindent\textbf{Running time breakdown.}
\label{sec:eval:time-decomposition} To identify the source of \name's speedup over naive offloading, we profile the runtime breakdowns for the Rubble and BigCity scenes on RTX 4090 testbed, as shown in Figure \ref{fig:eval_perf_decomp}. We have the following observations. (1) We observe significant communication and CPU Adam overheads in naive offloading, as discussed in \S\ref{bg:challenges-of-training-3dgs-on-a-consumer-gpu}. In both scenes, two overheads together account for more than 50\% of the training time. (2) In this figure, \name's pipeline running time takes into account both communication and computation, as their runtime cannot be well separated. And we observe that \name's pipeline runtime is notably shorter than the combined computation and communication duration in naive offloading. The overall acceleration results from overlapping communication with computation, along with reduced communication volume by transferring only in-frustum Gaussians. We also observe that \name's pipeline time, which includes both computation and communication, is only marginally longer than the naive offloading's computation-only time. This indicates that \name's  communication overhead on top of computation is very small. 
(3) The effect of overlapping CPU Adam differs depending on the scenes. Figure \ref{fig:eval_perf_decomp} illustrates that \name reduces CPU Adam latency more in Rubble than in BigCity through overlapping. 
This disparity arises because, first, CPU Adam requires more time in BigCity due to the increased number of Gaussians. Second, the lower resolution of BigCity means that its computation time is shorter than that of Rubble (see Figure \ref{fig:eval-throughput-compare-naive-offloading}), thus making it harder for CPU Adam to overlap. The lesser overlap of CPU Adam in BigCity explains why \name does not achieve the greatest speedup compared to naive offloading in BigCity, despite its lowest $\rho$ among all scenes. (4) Lastly, Figure~\ref{fig:eval_perf_decomp} shows that the scheduling overheads in \name involved in determining in-frustum Gaussian indices and computing the microbatch order based on TSP are marginal.

\begin{table}[t]
  \centering
  \scriptsize
  \begin{tabular}{l p{0.66\linewidth}}
    \toprule
    \textbf{Strategy} & \textbf{Description} \\ \midrule
    Random Order & Shuffle views uniformly at random. \\[2pt]
    Camera Order & Sort views by their camera-center coordinate along the scene’s principal axis.
    \\ \midrule
    GS Count Order & Sort descending by the number of Gaussians visible in each view. Prioritizing views with more Gaussians allows CPU Adam to update more Gaussians earlier, reducing its overhead. \\[2pt]
    TSP Order (\name) & Use TSP to find an order which maximizes Gaussian overlap between successive views. \\ 
    \bottomrule
  \end{tabular}
  \caption{\revise{Ordering strategies evaluated in our ablation study. The ``Random Order'' and ``Camera Order'' are straightforward; while both ``GS Count Order'' and ``TSP Order'' rely on view-Gaussian visibility information and thus require additional processing. 
  }   }
  \label{tab:ordering-strategies}
\end{table}

\begin{figure}[t]
    \centering
    \includegraphics[width=\linewidth]{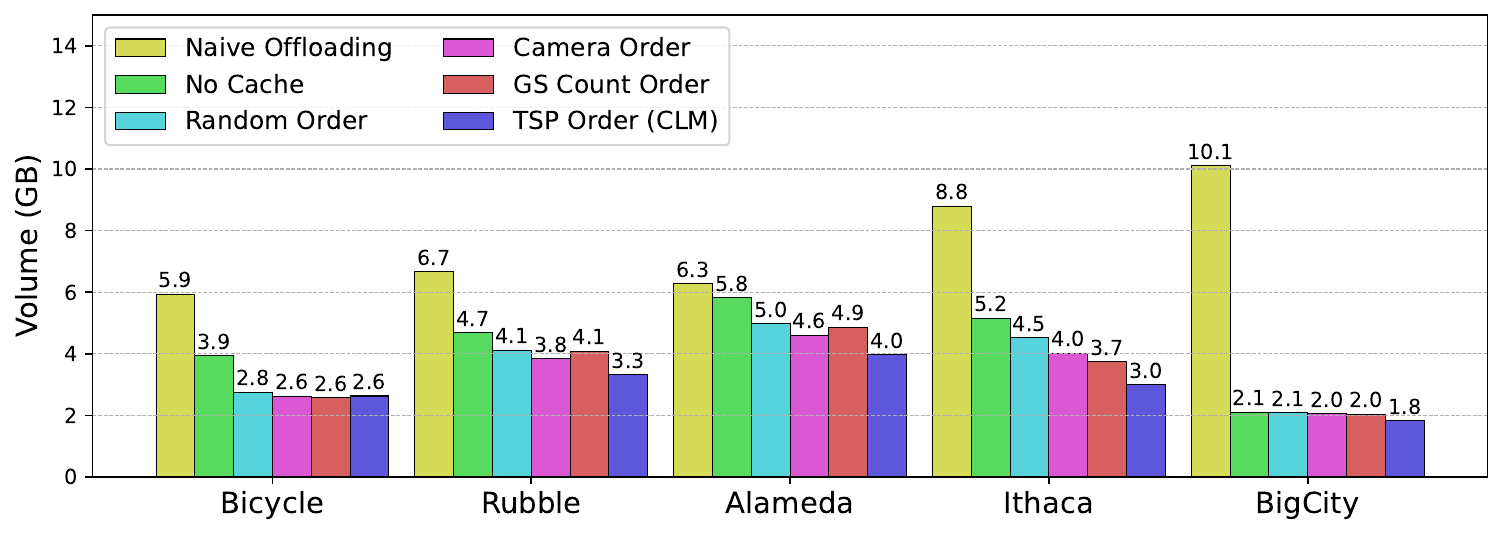}
    \caption{\revise{Average communication volume as measured by bytes transferred from CPU to GPU per training batch}. ``Naive Offloading'' denotes the communication volume for naive offloading without any optimization. ``No Cache'' denotes the volume for \name without Gaussian Caching and Order Optimization. \revise{The remaining four correspond to the ordering strategies detailed in Table \ref{tab:ordering-strategies}. } 
    }
    \label{fig:eval_CompVol}
\end{figure}

\noindent\textbf{Communication volume reduction.}
To better understand the communication overhead of \name, we collect the average communication volume per training batch, as shown in Figure \ref{fig:eval_CompVol}. 
\revise{The experiments for each scene use the maximum model size of naive offloading from Figure~\ref{fig:eval-max-model-size_4090}. We compare \name against naive offloading, and conduct ablation studies. 
Specifically, we evaluate a \name variant without Gaussian Caching (No Cache), as well as three variants with Gaussian Caching enabled, each of which uses a different ordering strategy: ``Random Order'', ``Camera Order'' and ``GS Count Order'', as described in Table~\ref{tab:ordering-strategies}. 
In contrast, \name uses TSP order (see Section \ref{sec:design:order-optimization}). These ablations help clarify how both caching and rendering order affect communication volume during training. We report the size of parameters transferred from CPU to GPU memories in GB. } 

\revise{Figure \ref{fig:eval_CompVol} shows that \name consistently decreases communication by 37\% (Alameda) to 82\% (BigCity) over naive offloading. 
In BigCity, the technique of selectively loading in-frustum Gaussians by itself significantly decreases communication (79\%), whereas Gaussian Caching offers small benefit, i.e. \name has 12\% additional reduction over ``No Cache''. This is because BigCity has a very low $\rho$ (\S\ref{fig:sparsity_cdf}), resulting in fewer Gaussians shared between two images for caching. Conversely, in scenes where each image contains a greater proportion of all Gaussians, like Bicycle, the use of Gaussian Caching yields more significant benefits, i.e. \name has 33\% additional reduction over ``No Cache''. We also observe that the TSP order consistently results in the lowest communication volume by maximizing cache reuse across microbatches. The greatest reduction is seen on the Ithaca scene --- 34\% lower than ``Random Order'', 25\% lower than ``GS Count Order'', and 19\% lower than ``Camera Order''. }

\begin{table}[h]
\centering
\scriptsize
\begin{subtable}[t]{0.5\textwidth}
    \centering
    \begin{tabular}{lccccc}
        \toprule
        \textbf{Method} & \textbf{Bicycle} & \textbf{Rubble} & \textbf{Alameda} & \textbf{Ithaca} & \textbf{BigCity} \\
        \midrule
        Random Order & 3.95 & 6.23 & 7.52 & 12.36 & 40.89 \\
        Camera Order & 3.94 & 6.27 & 7.58 & 12.67 & \textbf{40.94} \\
        GS Count Order & \textbf{4.06} & \textbf{6.65} & 8.01 & 12.50 & 40.80 \\
        TSP Order & 4.03 & 6.64 & \textbf{8.24} & \textbf{12.77} & 40.74 \\
        \bottomrule
    \end{tabular}
    \caption{Training throughput (img/s)}
    \label{tab:ablation_end2end}
\end{subtable}
\hfill
\begin{subtable}[t]{0.5\textwidth}
    \centering
    \begin{tabular}{lccccc}
        \toprule
        \textbf{Method} & \textbf{Bicycle} & \textbf{Rubble} & \textbf{Alameda} & \textbf{Ithaca} & \textbf{BigCity} \\
        \midrule
        Random Order & 147.457 & 137.27 & 178.59 & \textbf{208.34} & 21.85 \\
        Camera Order & 150.831 & 144.87 & 171.32 & 305.71 & \textbf{21.05} \\
        GS Count Order & \textbf{119.383} & \textbf{81.27} & \textbf{141.91} & 268.62 & 21.93 \\
        TSP Order & 147.457 & 127.11 & 186.75 & 406.79 & 23.36 \\
        \bottomrule
    \end{tabular}
    \caption{CPU Adam trailing time (ms) }
    \label{tab:ablation_cpuadam}
\end{subtable}
\caption{\revise{Average training throughput and CPU Adam trailing time under different ordering strategies (see Table~\ref{tab:ordering-strategies}). The  trailing time is calculated as the time spent by CPU Adam after the last gradient has been transferred from the GPU to the CPU. The more sophisticated strategies, ``TSP Order'' and ``GS Count Order'', deliver the highest end-to-end throughput. ``GS Count Order'' incurs the least CPU Adam trailing time; while Figure~\ref{fig:eval_CompVol} shows that ``TSP Order'' achieves the greatest reduction in communication volume. } 
} 
\end{table}

\revise{\noindent\textbf{Effectiveness of different ordering-strategies.} 
In Table~\ref{tab:ablation_end2end}, we compare the average training throughput (measured in processed images per second) across four ordering strategies in Table \ref{tab:ordering-strategies}: ``Random Order'', ``Camera Order'', ``GS Count Order'' and ``TSP order''. The experiments for each scene use the maximum model size supported by naive offloading from Figure~\ref{fig:eval-max-model-size_4090} on the 4090 testbed. 
We also report the corresponding communication volumes in Figure~\ref{fig:eval_CompVol} and CPU Adam ``trailing time'' 
in Table~\ref{tab:ablation_cpuadam}. We calculate ``trailing time'' as the time from when the last gradients are transferred to CPU memory to when CPU Adam finishes for the batch.  
Overall, smart reordering consistently improves training throughput over the default ``Random Order'', with the most significant gain observed on the Alameda scene---achieving a 10\% speedup. In contrast, BigCity shows minimal variation across orders in terms of communication volume, CPU Adam trailing time, and thus overall throughput. 
Among the strategies, ``TSP Order'' and ``GS Count Order'' achieve the highest throughput. ``TSP Order'' consistently minimizes communication volume, while ``GS Count Order'' reduces CPU Adam trailing time by rendering images that use more Gaussians earlier. 
An non-intuitive observation is that, for the Ithaca scene, the naive ``Random Order'' exhibits lower CPU Adam trailing time than ``GS Count Order''. 
This is because its communication is significantly slower than the other strategies (see Figure~\ref{fig:eval_CompVol}). The slower transfer allows more opportunity to overlap with CPU-side computation, thereby reducing its trailing time.}

\begin{figure*}[t]
    \centering
    \includegraphics[width=0.8\linewidth]{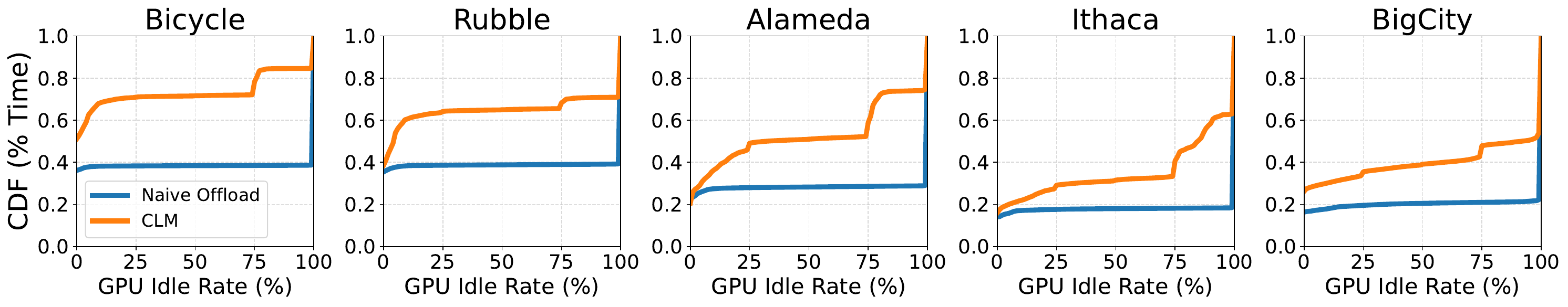}
    \caption{\revise{CDF of GPU Idle Rate (100 - \texttt{SMs Active}) Across Scenes for \name vs. Naive Offloading. Higher curves indicate more time spent at lower idle rates, reflecting better GPU utilization. \name consistently achieves higher GPU utilization across all scenes. } 
    } 
    \label{fig:eval_utilization_smactive_cdf}
\end{figure*}

\subsection{\revise{Hardware Utilization}} 

\revise{
\noindent\textbf{GPU utilization.} 
We compare the GPU utilization of \name against naive offloading by profiling using Nsight Systems \cite{nsightsystem} on the RTX 4090. We collect the \texttt{SMs Active} metric at 10 kHz from Nsight Systems in the \texttt{GPU\_METRICS} table. The values range from 0 to 100 and reflects the percentage of SMs with active warps in flight. A value of 0 indicates that all SMs are idle. Figure~\ref{fig:eval_utilization_smactive_cdf} presents the Cumulative Distribution Function (CDF) of the GPU idle rate, computed as 100 - \texttt{SMs Active}. The x-axis denotes the idle rate, while the y-axis represents the percentage of time. The area under the curve corresponds to the expected value of \texttt{SMs Active} during training, reflecting average GPU utilization. For each scene, both \name and naive offloading are profiled for the same duration---spanning more than 100 batches. We use the maximum model size for naive offloading from Figure \ref{fig:eval-max-model-size_4090}. We observe that \name consistently achieves better GPU utilization, as indicated by higher curves. Additionally, scenes with higher resolution (e.g., Bicycle and Rubble) exhibit better utilization compared to lower-resolution scenes (e.g., Ithaca and BigCity), confirming the intuition that higher-resolution rendering is more computational intensive. 
}

\begin{table}[t]
    \centering
    \resizebox{\columnwidth}{!}{
    \small
    \begin{tabular}{lccccc}
        \toprule
        \textbf{Testbed} & \textbf{Bicycle} & \textbf{Rubble} & \textbf{Alameda} & \textbf{Ithaca} & \textbf{BigCity} \\
        \midrule
        RTX 2080 Ti & 6.0 & 8.2 & 8.4 & 13.4 & 17.5 \\
        RTX 4090 & 14.1 & 17.2 & 16.1 & 28.4 & 37.8 \\
        \bottomrule
    \end{tabular}
    }
    \caption{\revise{Pinned memory usage (in GB) of \name for each scene using the maximum model size reported in Figure~\ref{fig:eval-max-model-size}.} } 
    \label{tab:pinned-memory-usage}
\end{table}

\revise{
\noindent\textbf{Pinned memory usage. } 
Pinned memory is a limited resource, so we report \name’s usage on the two testbeds in Table~\ref{tab:pinned-memory-usage}, using the corresponding maximum model sizes shown in Figure~\ref{fig:eval-max-model-size}. Even for the largest BigCity model, pinned memory usage peaks at 17.5 GB on the RTX 2080 Ti testbed and 37.8 GB on the RTX 4090---under 10\% of the RTX 2080 Ti testbed's 256 GB RAM and 30\% of the RTX 4090 testbed’s 128 GB RAM, respectively. This efficiency stems from pinning only parameter and gradient tensors (which require GPU DMA) in \name, while optimizer and auxiliary states remain unpinned. We observe no system performance degradation from this usage. 
} 

\revise{We further report CPU cores, PCIe bandwidth and GPU memory bandwidth utilization in Appendix~\ref{appendix:hardware-utilization}. }

%% file: 05_related_revision.tex
\section{Related Works}
\label{sec:related}

\paragraph{Current Approaches to Scaling 3DGS. } Several approaches have been suggested to decrease the GPU memory usage of 3DGS. First, systems like \cite{grendel,dogs,retinags} use the aggregate memory of multiple GPUs to distribute 3DGS training. However, the need for multiple GPUs and high-performance interconnects adds significant costs, putting these approaches out of reach for most users. 
Second, some approaches prune Gaussians ~\cite{lp3dgs,speedysplat, reduce3dgs,lightgaussian, minisplatting} that do not contribute significantly to the rendered image.
\revise{Although pruning methods are effective at reducing memory overheads, they can potentially hurt fidelity~\cite{lp3dgs,speedysplat} when Gaussians are pruned aggressively. Our approach is orthogonal, and does not affect fidelity.
Furthermore, for very large scenes, even a pruned model might not fit in a single-GPU's memory, making it necessary to combine pruning with approaches such as ours.} 
Furthermore, most of these approaches prune Gaussians after training, and thus cannot be used to scale 3DGS training. 
Third, divide-and-conquer methods \cite{citygaussian,vastgaussian,hierarchicalgaussians} partition scenes into smaller regions, process each partition in isolation, and finally reconstruct views by combining results from multiple partitions. But Gaussians at partition borders are likely to be used by multiple partitions, and in practice this leads to rendering inconsistencies at partition boundaries and lower reconstruction quality. Furthermore, these techniques are often more complex to use because of additional hyperparameters to tune.

\paragraph{CPU offloading in other ML training system.} As the size of ML models increases, the limitations of GPU memory have led to a greater use of CPU offloading techniques for ML training. For example, \cite{zerooffload, swapadvisor, vDNN} focus on DNN training in general, while \cite{offload-dlrm, cdlrm} specialize in recommendation systems, \cite{NeutronOrch, DGL-UVA} in GNNs, and \cite{llm-offload} in LLMs, all of which optimize for workloads distinct from 3DGS. Their focused workloads involve tensor operations, such as GEMM; whereas 3DGS is unique differentiable rendering pipeline. There are no studies that have successfully applied these offloading techniques to 3DGS training. In contrast, we demonstrate the feasibility of offloading in 3DGS and introduce tailored designs for improved efficiency. 
UGache \cite{UGache} optimizes the sparse access to CPU-based embedding tables, which is common in the offloaded training for the GNN and the recommendation system. Our Gaussian attributes in CPU memory akins to embedding table and is also accessed sparsely. However, UGache assumes that the embedding table is read-only, which is not applicable in 3DGS training. Moreover, UGache overlooks aspects such as spatial locality that are present in 3DGS, thereby missing some opportunities for optimization.

\paragraph{Other 3DGS Training System. } Recent studies have developed other hardware and software systems to improve 3DGS Training. 
GauSPU \cite{GauSPU}, GScore \cite{gscore} and MetaSapiens \cite{metasapiens}, ACR \cite{ACR} design hardware accelerators specifically for the rendering pipeline in 3DGS training. However, these accelerators are mainly optimized for speed or energy efficiency. Unlike these, \name emphasizes optimizing GPU memory efficiency. Additionally, our offloading methods may complement these accelerators by allowing a novel view synthesis task to utilize them for speed improvement while using our techniques for extra memory capacity.

%% file: 06_conclusion_revision.tex
\section{\revise{Discussion and Future Work}}\label{sec:discussion}

\revise{
Finally, we discuss how \name can be generalized to other rendering methods and backends, and future directions that use spatial data structures to further aid with scaling.}

\revise{\noindent\textbf{Support for other rendering backends and methods. }
\name is backend-agnostic because it determines where to store data (offloading), how to transfer it (overlapping), and when to render each image (pipelining and ordering), without depending on the specific rendering procedure. This decoupling enables seamless integration with APIs such as Vulkan and alternative rendering approaches like ray tracing, without requiring changes to \name’s scheduling or offloading logic. 
Furthermore, the core pipeline naturally extends to a broader class of point-based differentiable rendering techniques, such as 2D Gaussian Splatting~\cite{2dgs} and 3D Convex Splatting~\cite{3dcs}, due to their similar reliance on sparse data access patterns induced by frustum culling. 
However, \name cannot generalize to non-point-based novel view synthesis methods, like NeRF\cite{nerf}. }

\revise{
\noindent\textbf{Integration of spatial data structures.} 
As scenes grow larger and more complex, the number of Gaussians increases significantly. Although naive frustum culling---iterating over every Gaussian---is not a yet bottleneck in our current evaluation,
it could eventually become one as its time complexity scales linearly with the number of Gaussians. Future work could explore integrating spatial acceleration structures, such as bounding volume hierarchies (BVHs), to organize Gaussians more efficiently and thereby improving frustum culling performance by skip non-intersected regions. } 

\revise{
\noindent\textbf{Portability to other GPUs.}
Our \name implementation relies on two CUDA features of NVIDIA GPUs: pinned memory for direct memory access (DMA) and multi-streaming to overlap data transfer with computation. Both pinned-memory DMA and multi-streaming are standard features in modern GPUs (e.g., in AMD ROCm\cite{rocm}), and thus do not fundamentally limit portability. 
}

\section{Conclusion}\label{sec:conclusion}
Our goal in designing \name was to allow 3DGS to be used with larger scenes, without needing to compromise on rendering quality or pay for multi-GPU training. We were able to meet our goals because of the inherent sparsity of 3DGS's computation and its memory access pattern, that allowed us to overlap GPU-CPU communication, GPU computation and CPU computation. 
Our approach does not depend on details of how views are rasterized or what kernels are used, and therefore it can be applied to other novel view synthesis or ML methods that exhibit similar computation and memory access patterns.

%% file: 07_appendix_revision.tex
\newpage

\section{Appendix}

\subsection{Search for TSP solution} 
\label{sec:appendix-tsp}

We formulate the training image reordering task into a TSP problem, in which each training order corresponds to a tour in the TSP instance. We implement a Stochastic Local Search with well-established greedy heuristics (2-opt and 3-opt \cite{2opt} in our case). Our algorithm starts with an initial feasible tour, and then iteratively improves the tour by applying local greedy swapping. Our initialized tour is as follows: starts from a randomly chosen city and repeatedly selects the nearest unvisited city as the next destination. In the initialized tour, the nodes are connected by edges to form a chain. During every iterative improvement, we select 2 or 3 edges and remove these edges and reconnect the segments in a new way that reduces the tour length. We perform swaps until no further improvement is found or an adjustable time limit is reached. In our experiments, we use 1 ms as the time limit which is empirically sufficient for us to find an optimal solution (compared to DP-based method). The impressive results could be attributed to the relatively small batch size (the number of nodes in TSP) and ours is a variant of the TSP known as the metric TSP, which is typically easier to address empirically. The metric TSP mandates that the distance function is symmetric and fulfills the triangle inequality, which the symmetric distance complies with.

\subsection{Dataset Preparation}
\label{appendix:datasets-prepare} 

3DGS training requires a camera pose for each image. The Bicycle\cite{mipnerf360}, Rubble\cite{meganerf}, Alameda\cite{zipnerf}, and Matrixcity \cite{matrixcity} datasets initially support novel view synthesis and come with camera poses. However, the Ithaca dataset, initially designed for autonomous driving, does not include camera poses, so we use colmap \cite{colmap} to generate camera poses and the point cloud (used for initializing Gaussians) ourselves. 

The matrixcity dataset \cite{matrixcity} comprises sub-scenes of varying sizes and offers two perspectives: aerial and street. For our evaluation, matrixcity BigCity refers to the largest one among the aerial views scenes. 

We conducted experiments on all scenes at their native resolutions, without image downsampling.

\subsection{Fragmentation decreases the available memory for accommodating Gaussians} 
\label{appendix:fragmentation}
Given most 3DGS training pipeline (and ours) are built on pytorch, another factor that limits available memory for training is memory fragmentation due to the Pytorch Cache Allocator. The PyTorch Cache Allocator manages GPU memory by maintaining a pool of allocated memory blocks to reduce the overhead of frequent memory allocations and deallocations. Although this approach improves speed in many scenarios, it can lead to fragmentation over time, especially in workloads with varying allocation sizes. 3DGS exactly exhibits varying activation states during training, and their model states are frequently densified and pruned, leading to substantial fragmentation challenges. This fragmentation further reduces available GPU memory, hindering the accommodation of additional Gaussian parameters.

\subsection{Additional Hardware Utilization} 
\label{appendix:hardware-utilization} 

\begin{table}[htbp]
\centering
\resizebox{\linewidth}{!}{
\begin{tabular}{lrcc}
\toprule
\textbf{Scene} & \textbf{Metric} & \textbf{Naive offloading (\%)} & \textbf{CLM (\%)} \\
\midrule
Bicycle & CPU Util & 18.68 & \textbf{68.98} \\
 & DRAM Read & 9.61 & \textbf{16.72} \\
 & DRAM Write & 7.64 & \textbf{12.55} \\
 & PCIe RX & 12.62 & \textbf{20.85} \\
 & PCIe TX & 14.18 & \textbf{14.20} \\
\midrule
Rubble & CPU Util & 21.47 & \textbf{64.83} \\
 & DRAM Read & 10.29 & \textbf{15.86} \\
 & DRAM Write & 7.79 & \textbf{11.52} \\
 & PCIe RX & 12.38 & \textbf{20.94} \\
 & PCIe TX & 13.59 & \textbf{14.09} \\
\midrule
Alameda & CPU Util & 22.64 & \textbf{76.80} \\
 & DRAM Read & 7.27 & \textbf{10.61} \\
 & DRAM Write & 6.03 & \textbf{8.34} \\
 & PCIe RX & 14.47 & \textbf{30.88} \\
 & PCIe TX & 16.07 & \textbf{20.53} \\
\midrule
Ithaca & CPU Util & 24.97 & \textbf{82.44} \\
 & DRAM Read & 8.01 & \textbf{12.37} \\
 & DRAM Write & 5.17 & \textbf{7.37} \\
 & PCIe RX & 16.78 & \textbf{17.61} \\
 & PCIe TX & \textbf{19.11} & 12.57 \\
\midrule
BigCity & CPU Util & 25.24 & \textbf{61.95} \\
 & DRAM Read & 8.84 & \textbf{16.14} \\
 & DRAM Write & 2.89 & \textbf{5.17} \\
 & PCIe RX & \textbf{15.37} & 10.13 \\
 & PCIe TX & \textbf{16.97} & 7.13 \\
\bottomrule
\end{tabular}
}
\caption{\revise{Hardware Utilization of \name and Naive Offloading Across Five Scenes on RTX 4090. 
\texttt{CPU Util} refers to CPU cores utilization. 
\texttt{DRAM Read/Write} indicate GPU memory bandwidth utilization. \texttt{PCIe RX} represents PCIe CPU-to-GPU direction utilization, and \texttt{PCIe TX} represents GPU-to-CPU direction utilization. All values are percentages of utilization, ranging from 0 to 100. In each row, the bold figure is the one with the higher utilization.} } 
\label{tab:performance_comparison_with_cpu} 
\end{table}

\revise{We additionally report the utilization of CPU cores, GPU DRAM bandwidth and PCIe bandwidth  for both \name and naive offloading across all scenes on RTX 4090, as shown in Table~\ref{tab:performance_comparison_with_cpu}. 
To obtain CPU utilization, we extract thread scheduling events from Nsight Systems’ \texttt{SCHED\_EVENTS} table, which logs timestamps for entering and leaving each CPU core. We calculate the percentage of time each core has a thread in flight, then average across all cores to obtain overall CPU utilization. Additionally, we collect other metrics at a sampling rate of 10 kHz from Nsight Systems' \texttt{GPU\_METRICS} table: \texttt{DRAM Read Bandwidth}, \texttt{DRAM Write Bandwidth}, \texttt{PCIe RX}, and \texttt{PCIe TX}. These metrics reflect the read and write bandwidth utilizations for both GPU Memory and PCIe, respectively. All utilization values are percentages, ranging from 0 to 100. }

\revise{For CPU core utilization, \name consistently achieves higher usage than naive offloading. This is because \name overlaps CPU-side Adam optimization---the primary CPU workload---with GPU computation and communication. In contrast, naive offloading leaves most CPU cores idle while the GPU is computing or transferring Gaussians between CPU and GPU memory. }

\revise{For DRAM bandwidth, \name consistently exhibits higher utilization than naive offloading. This is because both approaches perform the same amount of memory access (as the rendering operations are the same), but \name consistently runs faster, resulting in higher bandwidth utilization over time. }

\revise{For PCIe utilization, \name generally shows higher values than naive offloading, except in Ithaca’s \texttt{PCIe TX} and BigCity’s \texttt{PCIe RX} and \texttt{PCIe TX}. 
In these 3 cases, naive offloading transfers significantly more data than \name in each batch (see Table~\ref{fig:eval_CompVol}), leading to higher utilization. Notably, \name achieves higher PCIe utilization in most other cases despite transferring less data. We also observe that \texttt{PCIe RX} (CPU-to-GPU) utilization in \name is consistently higher than \texttt{PCIe TX} (GPU-to-CPU), due to gradient accumulation in \name: old gradients are loaded from CPU pinned memory to each CUDA kernel via DMA, summed with new gradients, and written back. This results in bidirectional PCIe usage during gradient offloading, whereas parameter loading is unidirectional from CPU to GPU. 
Lastly, overall PCIe utilization is low, because the CPU/GPU may be busy and not sending data, and the transfers may be too sparse to fully saturate the bandwidth even when PCIe is active. }